\documentclass[reqno]{amsart}
\usepackage[foot]{amsaddr}
\usepackage{arxiv_style}

\theoremstyle{plain}
\newtheorem{theorem}{Theorem}[section]
\newtheorem{corollary}[theorem]{Corollary}
\newtheorem{lemma}[theorem]{Lemma}
\newtheorem{proposition}[theorem]{Proposition}
\theoremstyle{definition}
\newtheorem{definition}[theorem]{Definition}
\newtheorem{remark}[theorem]{Remark}

\newcommand{\R}{\mathbb{R}}
\newcommand{\C}{\mathbb{C}}
\newcommand{\N}{\mathbb{N}}
\newcommand{\Z}{\mathbb{Z}}
\newcommand{\Npos}{\N_+}
\newcommand{\Rnonneg}{\R_{\ge0}}
\newcommand{\bbP}{\mathbb{P}}
\newcommand{\bbE}{\mathbb{E}}

\newcommand{\M}{\mathcal{M}}
\newcommand{\calK}{\mathcal{K}}
\newcommand{\calL}{\mathcal{L}}
\newcommand{\calR}{\mathcal{R}}
\newcommand{\calZ}{\mathcal{Z}}
\newcommand{\rmL}{\mathrm{L}}
\newcommand{\rmV}{\mathrm{V}}
\newcommand{\rmW}{\mathrm{W}}

\newcommand{\myvector}[1]{\mathbf{#1}}
\newcommand{\myvectorgreek}[1]{\boldsymbol{#1}}
\newcommand{\mymatrix}[1]{\mathbf{#1}}
\newcommand{\mymatrixgreek}[1]{\boldsymbol{#1}}

\newcommand{\va}{\myvector{a}}
\newcommand{\vb}{\myvector{b}}
\newcommand{\vc}{\myvector{c}}
\newcommand{\ve}{\myvector{e}}
\newcommand{\vh}{\myvector{h}}
\newcommand{\vq}{\myvector{q}}
\newcommand{\vs}{\myvector{s}}
\newcommand{\vu}{\myvector{u}}
\DeclareRobustCommand{\vv}{\myvector{v}}
\newcommand{\vw}{\myvector{w}}
\newcommand{\vx}{\myvector{x}}
\newcommand{\vy}{\myvector{y}}
\newcommand{\vz}{\myvector{z}}
\newcommand{\mA}{\mymatrix{A}}
\newcommand{\mM}{\mymatrix{M}}
\newcommand{\mV}{\mymatrix{V}}
\newcommand{\mW}{\mymatrix{W}}
\newcommand{\vepsilon}{\myvectorgreek{\epsilon}}
\newcommand{\vzeta}{\myvectorgreek{\zeta}}
\newcommand{\veta}{\myvectorgreek{\eta}}
\newcommand{\vtheta}{\myvectorgreek{\theta}}
\newcommand{\vmu}{\myvectorgreek{\mu}}
\newcommand{\mTheta}{\mymatrixgreek{\Theta}}

\newcommand{\defas}{\coloneqq}
\newcommand{\normsmall}[1]{\|{#1}\|}
\newcommand{\norm}[1]{\left\|{#1}\right\|}
\newcommand{\diag}{\mathrm{diag}}
\newcommand{\D}{\mathrm{D}}
\newcommand{\Linf}{\mathrm{L}^\infty}
\newcommand{\Lp}{\mathrm{L}^p}
\newcommand{\Lone}{\mathrm{L}^1}

\newcommand{\Symgn}{S_n}
\newcommand{\rstar}{r_\star}
\newcommand{\jstar}{j_\star}
\newcommand{\real}{\rho}
\newcommand{\intran}[2]{\{#1, \dots, #2\}}
\newcommand{\coef}{\mathrm{coef}}

\title{Irreducible shallow tanh neural networks are a manifold}

\author[I.~Bioli]{Ivan Bioli$^{\dagger,\ddagger}$}
\email{ivan.bioli@unipv.it}

\author[I.~C.~Gonnella]{Isabella Carla Gonnella$^{\S}$}
\email{igonnell@sissa.it}

\author[C.~Marcati]{Carlo Marcati$^{\mathparagraph}$}
\email{carlo.marcati@univ-lyon1.fr}

\author[F.~Pichi]{Federico Pichi$^{\S}$}
\email{fpichi@sissa.it} 

\address{$^{\dagger}$Department of Mathematics ``F.~Casorati'', University of Pavia, 27100 Pavia, Italy}
\address{$^{\ddagger}$Department of Civil Engineering and Architecture, University of Pavia, 27100 Pavia, Italy}
\address{$^{\S}$MathLab, Mathematics Area, SISSA, 34136 Trieste, Italy}
\address{$^{\mathparagraph}$Université Lyon 1, Centrale Lyon, INSA Lyon, Université Jean Monnet, CNRS, ICJ UMR5208, 69622 Villeurbanne, France}

\thanks{The authors thank Giancarlo Sangalli for comments on an earlier
  version of the
    manuscript and for fruitful discussions on the topic. IB, ICG and FP are
    members of the Gruppo Nazionale Calcolo Scientifico -- Istituto Nazionale
    di Alta Matematica (GNCS-INdAM). The work of IB has received funding from
    the European Union's Horizon Europe research and innovation programme under
    the Marie Sk\l{}odowska-Curie Action MSCA-DN-101119556 (IN-DEEP). CM
    acknowledges the support of the French National Research Agency (ANR)
    through grant ANR-24-CPJ1-0099-01.}

\begin{document}

\begin{abstract}
    Neural networks (NNs) have been increasingly used as function approximators%
, motivating the study of the geometry of their function spaces.
We study the set of functions represented by shallow neural networks with hyperbolic tangent activation, fixed input and output dimensions, and $n$ hidden neurons.
In particular, we prove that the subset represented by irreducible networks in the sense of Sussmann (\emph{Neural Networks}, 1992), i.e., networks with no redundant neurons, is a smooth embedded submanifold of the Sobolev spaces $\rmW^{\ell,p}$ on bounded sets for any $\ell\in\N$ and $p\in[1,+\infty]$.
The key step is to show that if a sequence of networks converges in $\rmW^{\ell,p}$ to an irreducible network with the same number of hidden neurons, then the parameters of the approximating networks remain bounded.
Finally, we
show
that this manifold has exactly $n+1$
connected components when both the input and output are scalar, and is connected
otherwise.

    \bigskip\noindent
    \textsc{Keywords.} Neural Networks, Neuromanifold, Embedded submanifolds,
    Differential geometry

    \smallskip\noindent
    \textsc{MSC 2020.} 58D15 (primary); 68T07, 46E35 (secondary)
\end{abstract}
\maketitle


\section{Introduction and main result}
The term \emph{neural network manifold} (or \emph{neuromanifold}) has been
widely used in the literature to denote the set of functions that a neural
network with a fixed architecture can represent as its parameters vary
\cite{amari_singularities_2006,elisenda_grigsby_functional_2025}. However, the word
\emph{manifold}
has been used in this context in a loose sense, without rigorous
justification, merely to indicate that neural networks with a fixed architecture
form a ``smoothly varying'' set of functions. In this paper, we show that this
terminology can be made rigorous for shallow networks with the hyperbolic
tangent activation.

We consider shallow tanh networks with input dimension $d$, output dimension
$d'$, and $n$ hidden neurons, defined on any nonempty, open and bounded set
$\Omega \subset \R^d$. For every $\ell\in\N$ and $p\in[1,+\infty]$,
\Cref{thm:main} establishes that the set of irreducible networks is a smooth
embedded submanifold of $\rmW^{\ell,p}(\Omega,\R^{d'})$ of dimension
$nd+n+nd'+d'$. Here, irreducibility is in the sense of
\cite{sussmann_uniqueness_1992}, i.e., no neuron has incoming or outgoing
weights that are all zero, and no two neurons have the same incoming weights and
bias up to a sign change. In other words, no neuron is redundant.

Our proof has three main ingredients: (i) Irreducibility allows us to identify
each network with its parameters, up to the symmetries of the network, i.e.,
sign changes and permutations of the neurons. (ii) The realization map, which
maps the parameters to the function they represent, has a full-rank injective
differential. Hence, the realization map is an immersion, and the tangent space
has the same dimension at every point. (iii) The realization map, restricted to
irreducible parameters up to symmetries, is continuous with a continuous
inverse. This last step is the key one in our proof. We obtain it from the inverse stability
of the realization map, which relates the convergence of networks, as functions,
to the behavior of their parameters. Specifically, we prove in
\Cref{thm:boundedness} that if a sequence of shallow $\tanh$ networks converges
in $\Lp$ to an irreducible network with the same number of neurons, then the
parameters of the sequence remain bounded.

Finally, \Cref{thm:connectedness} shows that the manifold has exactly $n+1$
connected components if both the input and the output are scalar, and that it is
connected otherwise. In the scalar case, we fully characterize the connected
components: each of them is determined by the number of neurons whose input and
output weights have the same sign.

\subsection{Main result}
Throughout, let $d\in\N_+ \coloneqq \{1, 2, \dots\}$ and let $\Omega\subset\R^d$ be a nonempty open
and bounded set.

\begin{definition}
    \label{def:tanh_nn}
    Let $d, d', n \in \N_+$. A \emph{shallow tanh neural network} with input
    dimension $d$, output dimension $d'$ and hidden dimension $n$ is a function
    $u_{\vtheta}: \Omega \subset\R^d\to \R^{d'}$ of the form
    \begin{equation}
        \label{eq:tanh_nn}
        u_{\vtheta}(\vx) \coloneqq \mA \cdot \tanh(\mW \cdot \vx + \vb) + \vs = \sum_{i=1}^{n} \tanh(\vw_i^\top \vx + b_i) \,\va_i + \vs,\qquad \vx\in\Omega,
    \end{equation}
    where $\tanh$ is applied componentwise. The parameters of the network are
    $\mW \in \R^{n \times d}$ with rows $\vw_1, \dots, \vw_n \in \R^d$, $\mA \in
        \R^{d' \times n}$ with columns $\va_1, \dots, \va_n \in \R^{d'}$, $\vb \in
        \R^n$, and $\vs \in \R^{d'}$. We stack them vertically into the parameter vector
    \[
        \vtheta \coloneqq [\vw_1;\dots;\vw_n;\vb;
            \va_1;\dots;\va_n;\vs] \in \R^N,
        \qquad \text{where}\quad N = n d + n + d' n + d'.
    \]
    We use the same notation for the extension of $u_{\vtheta}$ to the whole
    domain $\R^d$ given by the formula in \eqref{eq:tanh_nn}.
\end{definition}
We give the definition of irreducible neural networks, see
\cite{sussmann_uniqueness_1992}.
\begin{definition}
    \label{def:nondegenerate_tanh_nn}
    With the notation of \Cref{def:tanh_nn}, we say that the parameters
    $\vtheta$ are \emph{irreducible}, and that $u_{\vtheta}$ is an
    \emph{irreducible network}, if the following holds:
    \begin{enumerate}[(i)]
        \item \label{def:item:nondeg_1} $\vw_i \neq \mathbf{0}$ for all $i = 1, \dots, n$,
        \item \label{def:item:nondeg_2} $(\vw_i, b_i) \neq \pm (\vw_j, b_j)$ for all $i, j = 1, \dots, n$, with $i \neq j$,
        \item \label{def:item:nondeg_3} $\va_i \neq \mathbf{0}$ for all $i = 1,\dots, n$,
    \end{enumerate}
    and we denote the set of irreducible parameters by $\mTheta$, defined as
    \begin{equation}
        \label{eq:nondeg_params}
        \mTheta \coloneqq \Big\{ \vtheta \in \R^N : \vtheta \text{ satisfies \eqref{def:item:nondeg_1}, \eqref{def:item:nondeg_2}, \eqref{def:item:nondeg_3}}\Big\}.
    \end{equation}
\end{definition}

\begin{remark}
    The neural networks in \Cref{def:nondegenerate_tanh_nn} are said to be
    \emph{irreducible} because conditions \eqref{def:item:nondeg_1},
    \eqref{def:item:nondeg_2}, and \eqref{def:item:nondeg_3} ensure that there is no
    neuron that can be removed or merged with another while preserving
    the function represented by the network.
    Indeed, condition
    \eqref{def:item:nondeg_1} ensures that the $i$-th hidden neuron is not
    constant, condition \eqref{def:item:nondeg_2} ensures that no two hidden
    neurons are the same up to a sign change, and condition
    \eqref{def:item:nondeg_3} ensures that all hidden neurons contribute to the
    output of the network. We will further discuss this concept of redundancy in
    \Cref{sec:redundancy}.
\end{remark}

\begin{definition}\label{def:realization_map}
    With the same notation as in \Cref{def:tanh_nn}, and for $\ell\in\N$ and $p\in[1,+\infty]$, we define the
    \emph{realization map} for shallow tanh neural networks $\real:\R^N \to
        \rmW^{\ell,p}(\Omega, \R^{d'})$ as $\real(\vtheta) \coloneqq u_{\vtheta}$.
\end{definition}
We now present the three main results of this paper. The first establishes the
manifold structure of the set of irreducible shallow tanh neural networks,
which is our final goal. The second, concerning the boundedness of the
networks, is the key intermediate result exploited in the proof of the first.
The third describes the connected components of the manifold.
\begin{restatable}[Irreducible shallow tanh networks are a manifold]{theorem}{mainthm}
    \label{thm:main}
    Let $\Omega \subset \R^d$ be a nonempty open and bounded set, let $p \in [1,
            +\infty]$, let $\ell\in\N$, and let $d, d', n \in \N_+$. The set of
    irreducible shallow tanh neural networks with input dimension $d$,
    output dimension $d'$, and hidden dimension $n$, defined as
    \begin{equation}
        \M = \left\{ u_{\vtheta}(\vx) = \sum_{i=1}^{n} \tanh(\vw_i^\top \vx + b_i) \,\va_i + \vs \,\,\Big| \,\, \eqref{def:item:nondeg_1},
        \eqref{def:item:nondeg_2}, \eqref{def:item:nondeg_3} \text{ in \Cref{def:nondegenerate_tanh_nn} hold} \right\}
    \end{equation}
    is a smooth embedded
    submanifold of $\rmW^{\ell,p}(\Omega, \R^{d'})$ of dimension $N = n d +
        n + d' n + d'$.
\end{restatable}

The main difficulty in proving \Cref{thm:main} is to show that the realization map is a topological embedding.
The following theorem provides the essential inverse stability of the realization map: convergence
in $\Lp$ to an irreducible network with the same number of neurons forces
the parameters to remain bounded.
\begin{restatable}[Boundedness at irreducible networks]{theorem}{boundedness}
    \label{thm:boundedness}
    Let $\Omega \subset \R^d$ be nonempty, open and bounded, let $p \in [1,
            +\infty]$, and let $d, d', n \in \N_+$. Let $\vtheta \in \mTheta$
    and let $\{\vtheta_m\}_{m \in \N} \subset \R^N$ be an arbitrary sequence
    such that $u_{\vtheta_m} \to u_{\vtheta}$ in $\Lp(\Omega, \R^{d'})$. Then
    $\{\vtheta_m\}$ is bounded in $\R^N$.
\end{restatable}
\begin{remark}
    \label{rem:properness_Lp}
    \Cref{thm:boundedness} requires only convergence in $\Lp$, so it applies
    in particular to $\rmW^{\ell,p}$-convergent sequences for every
    $\ell\in\N$. Moreover, irreducibility is required only of the limit
    parameters $\vtheta$: the sequence parameters $\vtheta_m$ may violate any of conditions
    \eqref{def:item:nondeg_1}--\eqref{def:item:nondeg_3}. The resulting statement
    thus concerns arbitrary parameter sequences whose networks converge to an
    \emph{irreducible} network with the same number of neurons. In particular, we
    do not claim global properness of $\real$ as a map from $\R^N$ to
    $\Lp(\Omega, \R^{d''})$. There are indeed sequences of diverging parameters
    whose networks converge to the zero function \cite{petersen_topological_2021}.
\end{remark}

Finally, the following theorem describes the connected components of $\M$.
\begin{restatable}[Connected components of irreducible networks]{theorem}{connectedness}
    \label{thm:connectedness}
    Under the assumptions of \Cref{thm:main}, let $\M$ be the manifold of
    irreducible shallow tanh neural networks. Then, the following hold:
    \begin{enumerate}[(i)]
        \item If $d \ge 2$ or $d' \ge 2$, then $\M$ is connected.
        \item If $d = d' = 1$, then $\M$ has exactly
              $n+1$ connected components, namely
              \[
                  \M_k \coloneqq \bigl\{ u_{\vtheta} : \vtheta \in \mTheta,\
                  \#\{ i : w_i \cdot a_i > 0 \} = k \bigr\},
                  \qquad k = 0, \dots, n.
              \]
    \end{enumerate}
\end{restatable}


\begin{remark}
    \label{rem:sigmoid}
    \Cref{thm:main,thm:boundedness,thm:connectedness} carry over directly to
    the logistic sigmoid $\sigma(t)=1/(1+e^{-t})$. Indeed, the identity
    $\sigma(t)=(1+\tanh(t/2))/2$ gives
    \[
        \sum_{i=1}^n\sigma(\vw_i^\top\vx+b_i)\,\va_i+\vs
        =\sum_{i=1}^n\tanh\Bigl(\frac{\vw_i^\top}{2}\vx+\frac{b_i}{2}\Bigr)
        \frac{\va_i}{2}+\vs+\frac12\sum_{i=1}^n\va_i,
    \]
    and the corresponding change of parameters is a linear bijection of
    $\R^N$ that maps $\mTheta$ onto itself.
\end{remark}

\subsection{Related work and context}
In this work, we study the set of functions represented by shallow $\tanh$
neural networks with a fixed architecture as a subset of Sobolev spaces. Our
interest in this question comes from the optimization of neural networks in
the context of PDEs and variational problems, where neural networks are
increasingly used as function approximators. Novel optimization methods, such
as Natural Gradient Descent (NGD)
\cite{bioli_accelerating_2025,muller_position_2024,nurbekyan_efficient_2023},
are being developed specifically to exploit the underlying variational
structure of these problems. The geometry of $\M$ in \Cref{thm:main} is
induced by the $\rmW^{\ell,p}(\Omega)$-norm, which is a natural choice in such
settings. For $p=2$, the restriction of this norm to the tangent spaces
defines a Riemannian metric, which makes $\M$ a Riemannian manifold.
Therefore, NGD on the set of irreducible shallow $\tanh$ neural networks
becomes Riemannian gradient descent on a smooth manifold, and the theory of
optimization on manifolds
\cite{boumal_introduction_2023,absil_optimization_2008} could be applied to
study its convergence properties.
The singularities, i.e., the reducible parameters,
form a set of measure zero, but can still affect the training
dynamics \cite{amari_singularities_2006}.

The study of the topological and geometric properties of sets of neural
network realizations has been initiated and developed in
\cite{petersen_topological_2021,mahan2021nonclosedness,lindsey2026geometry,martinet2026closedness}, which show that these
sets are in general not closed, both in $\Lp$ and in Sobolev spaces. Earlier
related work on the geometry of neural networks focused instead on the
parameter space, proving that the Fisher information metric
\cite{amari_natural_1998} degenerates at singularities caused by redundant
hidden neurons \cite{amari_singularities_2006}.

Proving that the set $\M$ of irreducible networks is an embedded submanifold
of the Sobolev space $\rmW^{\ell,p}(\Omega)$ amounts to establishing three
properties of the realization map, modulo the symmetries of the parameters:
injectivity, injectivity of the differential, and continuity of the inverse.
The first property is the identifiability of $\tanh$ networks established in
\cite{sussmann_uniqueness_1992,fefferman1994} and later extended to more
general settings in \cite{vlacic_affine_2021,vlacic_neural_2022}. Related
questions concerning parameter symmetries, identifiability, and local
identifiability from a finite amount of data have also been studied for ReLU
networks~\cite{stock2023embedding}.
The second property
is closely related to a result of
\cite{fukumizu_regularity_1996},
which
proved that the Fisher information matrix of a shallow sigmoidal network
is positive definite if and only if the network is irreducible. The third
property is
the main difficulty of our proof, since the realization map is not inverse
stable \cite{petersen_topological_2021}: close realizations need not admit
close parameters. \Cref{thm:boundedness} rules this out near irreducible
networks by showing that parameters whose realizations converge to an
irreducible network with the same number of neurons remain bounded.
Together with identifiability, it can therefore be seen as a qualitative,
local inverse-stability result for shallow $\tanh$ networks at irreducible
parameters. For shallow ReLU networks,
\cite{berner_degenerate_2019} showed that inverse stability fails in general,
but holds with respect to a Sobolev norm on a suitably restricted set of
parameters.
\subsection{Structure of the paper}
In \Cref{sec:redundancy}, we study the parameter symmetries of irreducible
networks and the quotient of $\mTheta$ by these symmetries, on which the
realization map is injective. In \Cref{sec:properness_proof}, we prove
\Cref{thm:boundedness}, first in one dimension, and then extend it to higher dimensions by reduction to line
slices. In \Cref{sec:proof}, we show that the
differential of $\real$ is injective at irreducible parameters and combine
these results to finally prove \Cref{thm:main} and \Cref{thm:connectedness}.
\subsection{Notation}
We denote by $\Npos = \{1, 2, \dots \}$ the set of positive integers, and $\N =
    \{0\}\cup \N_+$.
We write $\Rnonneg = \{x\in\R: x\ge0\}$.
On a set $\Omega$, $\Lp(\Omega)$ are the classical Lebesgue spaces for
$p\in [1, \infty]$,
and $\rmW^{\ell,p}(\Omega)$ are Sobolev spaces of order $\ell \in \N$.
For vector-valued functions, we write
\begin{equation*}
    \| f\|_{\rmL^p(\Omega)}^p = \int_{\Omega} \| f\|_{\ell^p}^p\;,\qquad
    \|f\|_{\rmL^\infty(\Omega)}  = \operatorname*{ess\,sup}_{x\in\Omega} \|f(x)\|_{\ell^\infty}.
\end{equation*}
For $a, b\in \N$ with $b < a$ and any sequence $(c_j)_j$, we assume that the
empty product and sum have values
$\prod_{j=a}^bc_j =1$ and $\sum_{j=a}^b c_j=0$.
For $q,k\in\N$ and $I\subset\R$, we denote by $\bbP_q(I, \R^k) = \bbP_q(I, \R)^k$
the space of $\R^k$-valued polynomials of degree at most $q$ defined on $I$.
For a vector-valued polynomial of degree $q\in \N$ which can be written in the
monomial basis as $Q(t)=  \sum_{a=0}^q\vc_a t^a$, we write
\begin{equation}
    \label{eq:hashnorm}
    \|Q\|_\coef = \max_{a\in\intran{0}{q}}\|\vc_a\|_{\ell^\infty},
\end{equation}
and $\deg Q = \max\{a: \vc_a \neq \mathbf{0}\}$. For a multi-index
$\alpha \in \N^d$ and $\vx\in \R^d$, we write  $\vx^\alpha = \prod_k
    x_k^{\alpha_k}$ and $|\alpha| = \sum_{k=1}^d\alpha_k$, and we equip
$\rmW^{\ell,p}(\Omega,\R^{d'})$ with the norm $\|f\|_{\rmW^{\ell,p}(\Omega)} =
    \sum_{|\alpha|\le\ell}\|\partial^\alpha f\|_{\Lp(\Omega)}$. Finally, for a
measurable set $A\subset\R^s$, we denote by $|A| = \calL^s(A)$ its
$s$-dimensional Lebesgue measure.

\section{Removing redundancy in neural network realizations}
\label{sec:redundancy}

In this section, we show that excluding the redundancies identified by
\Cref{def:nondegenerate_tanh_nn} yields representations with a minimal number
of hidden neurons, unique up to neuron permutations and sign changes.
We then construct the quotient of $\mTheta$ by these remaining symmetries,
on which the induced realization map is injective.
Such symmetries form the finite group
\[
    G = (\Z_2)^n \rtimes \Symgn,
\]
i.e., the semidirect product of $(\Z_2)^n$, which we identify with $\{-1,+1\}^n$ under componentwise multiplication, and
the symmetric group $\Symgn$ of permutations of $\{1,\dots,n\}$.
Thus, the elements of $G$ are the pairs $g =
    (\vepsilon, \sigma)$ with $\vepsilon = (\epsilon_1, \dots, \epsilon_n) \in
    \{-1,+1\}^n$ and $\sigma \in \Symgn$, with multiplication
\begin{equation}
    \label{eq:group_law}
    (\vepsilon, \sigma) \cdot (\vepsilon', \sigma')
    = \bigl( (\epsilon_i\, \epsilon'_{\sigma^{-1}(i)})_{i=1}^n,\; \sigma \circ \sigma' \bigr),
\end{equation}
identity element $(\mathbf 1, \mathrm{id})$, and inverse $(\vepsilon,
    \sigma)^{-1} = ((\epsilon_{\sigma(i)})_{i=1}^n, \sigma^{-1})$.

\begin{definition}[Action of $G$]
    \label{def:action}
    For $g=(\vepsilon,\sigma)\in G$ and $\vtheta\in\mTheta$, we define
    $g\cdot\vtheta\coloneqq\vtheta'$, where the elements of $\vtheta'$ are $\vs'=\vs$ and
    \[
        (\vw_i',b_i',\va_i')
        =\epsilon_i(\vw_{\sigma^{-1}(i)},b_{\sigma^{-1}(i)},\va_{\sigma^{-1}(i)})
        \quad \forall i=1,\dots,n.
    \]
\end{definition}
Changing the signs of the neurons and permuting them preserves conditions
\eqref{def:item:nondeg_1}--\eqref{def:item:nondeg_3}, so $g\cdot\vtheta\in\mTheta$,
and \eqref{eq:group_law} gives $(gh)\cdot\vtheta=g\cdot(h\cdot\vtheta)$; hence
this is a left action of $G$ on $\mTheta$. Moreover, since $\tanh$ is odd,
$\real(g\cdot\vtheta)=\real(\vtheta)$ for every $g\in G$.

\begin{definition}[Equivalent parameters]
    \label{def:equivalence}
    Two parameter vectors $\vtheta,\vtheta'\in\mTheta$ are \emph{equivalent},
    and we write $\vtheta\sim\vtheta'$, if they lie in the same orbit of the
    action, i.e., if $\vtheta'=g\cdot\vtheta$ for some $g\in G$.
\end{definition}
Since $G$ is a group, $\sim$ is an equivalence relation, and the equivalence
class of $\vtheta$ is its orbit $[\vtheta]\coloneqq\{g\cdot\vtheta : g\in G\}$. We
denote by $\mTheta/G$ the set of equivalence classes and by
$\pi:\mTheta\to\mTheta/G$, $\pi(\vtheta)=[\vtheta]$, the canonical projection.
Equivalent parameter vectors are called ``isomorphic'' in
\cite{fefferman1994}. By the above, equivalent parameters realize the same
function. To prove the converse, we use the following linear-independence
result, which will also be useful in the proof of \Cref{thm:main}.
\begin{restatable}[Generalized linear independence]{proposition}{genindep}
    \label{lem:generalized_independence}
    Let $U \subseteq \R^d$ be open and nonempty. Let $(\vw_1, b_1), \dots,
        (\vw_R, b_R) \in \R^d \times \R$ satisfy conditions
    \eqref{def:item:nondeg_1} and \eqref{def:item:nondeg_2} of
    \Cref{def:nondegenerate_tanh_nn};
    let $\vmu_1, \dots, \vmu_S \in \R^d \setminus \{\mathbf 0\}$
    be pairwise distinct; and let $J, A \in \N$. Then the functions
    \[
        \begin{aligned}
                   & \big\{ \vx^\alpha : |\alpha| \le A \big\}                                                                   \\
            \cup\  & \big\{ \vx^\alpha \tanh^{(j)}(\vw_r^\top \vx + b_r) : 1 \le r \le R,\ 0 \le j \le J,\ |\alpha| \le A \big\} \\
            \cup\  & \big\{ \vx^\alpha e^{\vmu_s^\top \vx} : 1 \le s \le S,\ |\alpha| \le A \big\}
        \end{aligned}
    \]
    are linearly independent over $\R$ as functions on $U$. 
\end{restatable}

We defer the proof of this statement to \Cref{sec:lin_indep}.

The following lemma
concerns the \emph{identifiability} of shallow $\tanh$ networks: irreducible
parameters realizing the same function have the same number of hidden neurons
and are equivalent in the sense of \Cref{def:equivalence}.

\begin{restatable}{lemma}{identifiability}
    \label{lem:identifiability}
    Let $\vtheta \in \mTheta$ be irreducible parameters with hidden dimension
    $n$, and let $\vtheta' \in \R^{N'}$ be irreducible parameters with hidden
    dimension $n'\in\N$, where $N' = n' d + n' + d' n' + d'$.
    If $u_{\vtheta}=u_{\vtheta'}$ on $\Omega$, then $n'=n$ and
    $\vtheta\sim\vtheta'$.
\end{restatable}
This result is known: it follows from \cite[Theorem 1]{sussmann_uniqueness_1992}
for scalar output ($d'=1$), and from \cite[Proposition 2]{vlacic_affine_2021}
for general output dimension $d'$. The only difference in the statements is that
both references assume that the two networks realize the same function on the
whole of $\R^d$, whereas here equality is only assumed on $\Omega$. This is not
a real restriction: since $\tanh$ is real analytic, the difference of the two
realizations is real analytic on $\R^d$, and a real analytic function vanishing
on a nonempty open subset of the connected set $\R^d$ vanishes identically. For
completeness, in \Cref{sec:apx_identifiability} we give an elementary proof that
relies only on \Cref{lem:generalized_independence}.

We record two immediate consequences of \Cref{lem:identifiability}: the first
characterizes the fibers of $\real$ on $\mTheta$, and the second establishes
that an irreducible network with $n$ hidden neurons cannot be represented using
fewer hidden neurons.

\begin{corollary}
    \label{cor:non_injectivity}
    Let $\vtheta, \vtheta' \in \mTheta$ be two irreducible parameters. Then
    $\real(\vtheta) = \real(\vtheta')$ if and only if $\vtheta \sim \vtheta'$.
\end{corollary}

\begin{corollary}
    \label{cor:not_equal_lower_dim}
    Let $\vtheta \in \mTheta$ and $\vtheta' \in \R^{N'}$ be as in
    \Cref{lem:identifiability}, with $n' < n$. Then $u_{\vtheta} \neq
        u_{\vtheta'}$.
\end{corollary}

Finally, with the following lemmas, we establish the properties of the action
of $G$ on $\mTheta$ and the smooth manifold structure of the quotient $\mTheta/G$.

\begin{lemma}
    \label{lem:action_properties}
    The action of $G = (\Z_2)^n \rtimes \Symgn$ on $\mTheta$ is smooth, free,
    proper and isometric.
\end{lemma}
\begin{proof}
    For a fixed $g\in G$, the map
    \begin{align*}
        \mTheta & \longrightarrow \mTheta
        \\
        \vtheta & \longmapsto g\cdot \vtheta
    \end{align*}
    is linear, hence smooth. Since $G$ is discrete, this implies that the
    map
    \begin{align*}
        G \times \mTheta     & \longrightarrow \mTheta
        \\
        (g, \,\,\,\,\vtheta) & \longmapsto g\cdot \vtheta
    \end{align*}
    is also smooth.

    To prove that the action is free, we need to show that if there exists
    $\vtheta\in\mTheta$ such that $g\cdot\vtheta = \vtheta$, then $g =
        (\vepsilon, \sigma)$ is the identity element of $G$. Assume that $g\cdot
        \vtheta = \vtheta$. Take any $i\in\{1,\dots,n\}$ and let $j =
        \sigma^{-1}(i)$. Since $g\cdot\vtheta = \vtheta$, we have $(\vw_i, b_i,
        \va_i) = \epsilon_i (\vw_j, b_j, \va_j)$; in particular $(\vw_i, b_i) =
        \pm(\vw_j, b_j)$. Irreducibility (condition \eqref{def:item:nondeg_2})
    excludes the signed coincidence $(\vw_i, b_i) = \pm(\vw_j, b_j)$ for $i \neq
        j$, so it must hold that $j = i$. Thus, $\sigma = \mathrm{id}$. Then $\vw_i
        = \epsilon_i \vw_i$ with $\vw_i \neq \mathbf 0$ forces $\epsilon_i = 1$ for
    all $i$, and hence $g$ is the identity element of $G$.

    Since $G$ is finite and discrete, the action is proper by \cite[Proposition
        21.5]{lee_introduction_2012}. Finally, it is clear that for any $g\in G$ the map
    $\vtheta\mapsto g\cdot \vtheta$ is an isometry.
\end{proof}

We note that the quotient $\mTheta/G$ admits
a unique smooth manifold structure and Riemannian metric for which the
projection $\pi$ is a smooth Riemannian
submersion. Indeed, $\mTheta$ is open in $\R^N$ and inherits its Euclidean
Riemannian metric, so this follows from \Cref{lem:action_properties} and
\cite[Corollary 2.29]{lee_introduction_2018}.
Moreover, since $\rho:\mTheta\to\rmV=\rmW^{\ell,p}(\Omega,\R^{d'})$ is constant
on each orbit, it induces a map $\tilde{\rho}:\mTheta/G\to\rmV$
satisfying $\tilde{\rho}([\vtheta])=\rho(\vtheta)$, i.e., the following diagram commutes:
\[
    \begin{tikzcd}
        \mTheta \arrow[r,"\pi"] \arrow[dr,swap,"\rho"] &
        \mTheta / G \arrow[d,"\tilde{\rho}"]
        \\
        & \rmV
    \end{tikzcd}
\]
The smoothness of $\rho$ established in \Cref{lem:partial_realization}
implies that $\tilde{\rho}$ is smooth by
\cite[Corollary 9.21]{boumal_introduction_2023}, while
\Cref{cor:non_injectivity} gives its injectivity. Finally, the following
lemma provides a natural metric inducing the quotient topology on
$\mTheta/G$, which we will use in the proof of \Cref{thm:main}.

\begin{lemma}[Quotient metric]
    \label{lem:quotient_metric}
    The function
    \[
        d_G\bigl([\vtheta], [\veta]\bigr) \coloneqq \min_{g \in G} \norm{\vtheta - g\cdot\veta},
        \qquad [\vtheta], [\veta] \in \mTheta / G,
    \]
    is a well-defined metric on $\mTheta/G$, and it induces the
    quotient topology.
\end{lemma}
\begin{proof}
    The minimum is attained because $G$ is finite. Since the action is
    isometric and linear, for every $g, h, k \in G$ we have
    \[
        \norm{h\cdot\vtheta - g\cdot(k\cdot\veta)}
        = \norm{\vtheta - (h^{-1}gk)\cdot\veta},
    \]
    so
    \[
        \min_{g\in G}\norm{h\cdot\vtheta - g\cdot(k\cdot\veta)}
        = \min_{g\in G}\norm{\vtheta - g\cdot\veta}:
    \]
    the definition does not depend on the chosen representatives. Symmetry follows from $\norm{\vtheta - g\cdot\veta}
        = \norm{g^{-1}\cdot\vtheta - \veta}$ and minimizing over $g$. For the
    triangle inequality, for any $g, h \in G$,
    \[
        \norm{\vtheta - (gh)\cdot\vzeta}
        \le \norm{\vtheta - g\cdot\veta} + \norm{g\cdot\veta - (gh)\cdot\vzeta}
        = \norm{\vtheta - g\cdot\veta} + \norm{\veta - h\cdot\vzeta},
    \]
    and minimizing over $g$ and $h$ gives
    \[
        d_G([\vtheta],[\vzeta]) \le d_G([\vtheta],[\veta]) + d_G([\veta],[\vzeta]).
    \]
    If
    $d_G([\vtheta],[\veta]) = 0$, then $\vtheta = g\cdot\veta$ for some $g$
    (the minimum being attained), so $[\vtheta] = [\veta]$.

    It remains to prove that $d_G$ induces the quotient topology. First,
    $d_G$-balls are open in the quotient topology: indeed
    \[
        \pi^{-1}\bigl(B_{d_G}([\vtheta], r)\bigr)
        = \{\veta \in \mTheta : \exists g \in G, \ \norm{\vtheta - g \cdot \veta} < r\}
        = \bigcup_{g\in G} B\bigl(g^{-1}\cdot\vtheta,\, r\bigr) \cap \mTheta
    \]
    is open in $\mTheta$. Conversely, let $V \subseteq \mTheta/G$ be
    open in the quotient topology and $[\vtheta] \in V$. Then $\pi^{-1}(V)$ is
    open in $\mTheta$ and contains $\vtheta$, so $B(\vtheta, r) \subseteq
        \pi^{-1}(V)$ for some $r > 0$ small enough that also $B(\vtheta, r)
        \subseteq \mTheta$. For any $[\veta]$ with $d_G([\vtheta],[\veta]) < r$
    there is $g$ with $g\cdot\veta \in B(\vtheta, r) \subseteq \pi^{-1}(V)$
    and $\pi(g\cdot\veta) = [\veta] \in V$; hence $B_{d_G}([\vtheta], r) \subseteq
        V$. Therefore the $d_G$-balls form a basis of the quotient topology.
\end{proof}

\section{Proof of the boundedness theorem}
\label{sec:properness_proof}
In this section we prove \Cref{thm:boundedness}. We first prove it for
one-dimensional inputs (\Cref{sec:properness_1d}), and then show how this
implies the general case, by reasoning coordinatewise in a well-chosen basis
of $\R^d$ (\Cref{sec:properness_multid}).

\subsection{The one-dimensional case}
\label{sec:properness_1d}
We prove the one-dimensional case, \Cref{lem:properness_interval}, following
the strategy below. Given networks $( g_m )_m$ converging in $\Lone$ on a bounded
interval to a network $g$, we extract a subsequence along which each weight and bias of
each neuron has a (finite or infinite) limit,
and we proceed as follows.
\begin{enumerate}[(1)]
    \item \emph{Clusters.} The neurons whose parameters converge to a finite weight-bias
          pair are grouped into clusters with a common (up to sign) limit. Using divided differences (\Cref{lem:properness_newton}),
          \Cref{lem:properness_cluster} expands the contribution of each
          cluster into derivatives of $\tanh$ at the common limit, with
          polynomial coefficients.
    \item \emph{Degenerate neurons.} Each of the remaining neurons has a
      subsequence that converges
          either to a constant or to a step function. On a subinterval that does
          not contain the
          jumps of the latter,
          there exists a subsequence, along which, if their contribution
          converges, it converges to an
          extended exponential sum (\Cref{lem:properness_elimination}).
    \item \emph{Local structure of the limit.} Combining the two previous
          steps, \Cref{prop:properness_local_structure} shows that, on a
          subinterval, the limit of a subsequence of $( g_m )_m$
          is the sum of an extended exponential sum and
          of derivatives of $\tanh$ at the limits of the clusters, with
          polynomial coefficients.
    \item \emph{Matching.} In \Cref{lem:properness_interval}, we compare this
          description of the limit  with the limiting network $g$ using the
          linear independence of \Cref{lem:generalized_independence}. Since
          each neuron must converge to a distinct target neuron, all parameters
          of $g$ are bounded.
\end{enumerate}

\begin{definition}[Hermite interpolant and divided differences]
    \label{def:hermite_interpolant}
    Let $h\in C^\infty(\R)$ and let $x_1,\ldots,\allowbreak x_k$ be real numbers, not
    necessarily distinct. For a node $z\in \{x_1,\ldots,x_k\}$,
    let $\mu_z = \#\{l: x_l=z\}$
    be the number of indices
    such that $x_l=z$. The \emph{Hermite interpolant} of $h$ at $x_1,\ldots,x_k$
    is the unique polynomial $\mathcal{I}$ of degree at most $k-1$ such that
    $\mathcal{I}^{(r)}(z)=h^{(r)}(z)$ for every node $z$ and $r\in\intran{0}{\mu_z-1}$
    \cite[Proposition~7]{deboor_divided_2005}. When the nodes are distinct, $\mathcal{I}$
    is the Lagrange interpolation polynomial. The \emph{divided difference}
    of $h$ at the nodes $\{x_1, \dots, x_k\}$, denoted
    $h[x_1,\ldots,x_k]$, is the coefficient of $x^{k-1}$ in $\mathcal{I}$; it is zero if
    $\mathcal{I}$ has degree less than $k-1$. By \cite[Proposition~7]{deboor_divided_2005}, the Hermite interpolant of $h$
    at the nodes $\{x_1, \dots, x_k\}$ has the Newton form
    \begin{equation}
        \label{eq:hermite_netwon}
        \mathcal{I}(x)=\sum_{q=0}^{k-1}h[x_1,\ldots,x_{q+1}]
        \prod_{r=1}^q(x-x_r).
    \end{equation}
\end{definition}

\begin{lemma}
    \label{lem:properness_newton}
    For $\kappa\in \N$, let $x_1,\ldots,x_\kappa$ be real numbers
    (not necessarily distinct), let
    $h\in C^\infty(\R)$, and let $h[x_1,\ldots,x_{q+1}]$, for all $0\le q<\kappa$, be
    the divided differences. Then, for any $\vc_l\in\R^{d'}$,
    \begin{equation}
        \label{eq:hermite_newton_vector}
        \sum_{l=1}^{\kappa}\vc_lh(x_l)=
        \sum_{q=0}^{\kappa-1}h[x_1,\ldots,x_{q+1}]
        \sum_{l=1}^{\kappa}\vc_l\prod_{r=1}^q(x_l-x_r).
    \end{equation}
    Moreover, for any $x^*\in\R$ and any $q\in \intran{0}{\kappa-1}$,
    \begin{equation}
        \label{eq:dd_approx}
        \left|h[x_1,\ldots,x_{q+1}]-\frac{h^{(q)}(x^*)}{q!}\right|
        \le\frac{\|h^{(q+1)}\|_{\Linf(\R)}}{q!}
        \max_{l\le q+1}|x_l-x^*|,
    \end{equation}
    whenever the derivative norm on the right is finite.
\end{lemma}
\begin{proof}
    Since $\mathcal{I}$ agrees with $h$ at every node, $h(x_l)=\mathcal{I}(x_l)$ for
    $l\in\intran{1}{\kappa}$. Hence, the first part of the statement follows from \eqref{eq:hermite_netwon}.

    To prove the second estimate, fix $q\in\intran{0}{\kappa-1}$ and let
    $a=\min_{l\le q+1}x_l$, $b=\max_{l\le q+1}x_l$. By the mean value
    property of divided differences
    \cite[eq.~(44)]{deboor_divided_2005}, there exists
    $\xi\in[a,b]$ with
    \begin{equation}
        \label{eq:properness_newton_rolle}
        h[x_1,\ldots,x_{q+1}]=\frac{h^{(q)}(\xi)}{q!}.
    \end{equation}
    Now let $x^*\in\R$. By
    \eqref{eq:properness_newton_rolle} and the mean value theorem
    applied to $h^{(q)}$, there is $\eta$ between $x^*$ and $\xi$ with
    \begin{align*}
        \left|h[x_1,\ldots,x_{q+1}]-\frac{h^{(q)}(x^*)}{q!}\right|
         & =\frac1{q!}\bigl|h^{(q)}(\xi)-h^{(q)}(x^*)\bigr|
        =\frac1{q!}\,|\xi-x^*|\,\bigl|h^{(q+1)}(\eta)\bigr|    \\
         & \le\frac{\|h^{(q+1)}\|_{\Linf(\R)}}{q!}\,|\xi-x^*|.
    \end{align*}
    Finally, $\xi\in[a,b]$ and $a,b$ are nodes, so
    $|\xi-x^*|\le\max(|a-x^*|,|b-x^*|)\le\max_{l\le q+1}|x_l-x^*|$.
\end{proof}

The next lemma provides an expansion for the combined contribution of
$\kappa$ neurons whose weights and biases $\{(w_l,b_l)\}_{l=1}^\kappa$ are close to a common
center $(w_*,b_*)$. This contribution is approximated by a sum of $\kappa$
terms, each consisting of a derivative $\tanh^{(q)}(w_*t+b_*)$ multiplied
by a polynomial of degree at most $q$, for $q=0,\dots,\kappa-1$.
A simple example illustrates how derivatives of $\tanh$ arise.
For $h\neq0$, consider the sum of the two neurons
$\frac1h\tanh((1+h)t)$ and $-\frac1h\tanh(t)$.
Taylor's theorem gives, for $|t|\le R$,
\[
    \Bigl|\frac{\tanh((1+h)t)-\tanh(t)}{h}-t\tanh'(t)\Bigr|
    \le\frac{|h|R^2}{2}\,\|\tanh''\|_{\Linf(\R)}.
\]
Thus, the sum is approximated by $t\tanh'(t)$, i.e., the product of a polynomial and a derivative of $\tanh$.
In the notation of \Cref{lem:properness_cluster} below, this example
corresponds to $d'=1$, $\kappa=2$, $(w_1,b_1)=(1+h,0)$,
$(w_2,b_2)=(1,0)$, $a_1=-a_2=1/h$, and $(w_*,b_*)=(1,0)$,
for which $N_0=0$, $N_1(t)=t$, and $\delta=R|h|$.

\begin{lemma}
    \label{lem:properness_cluster}
    Let $J\subset\R$ be a nonempty bounded open interval, let
    $R=\sup_{t\in J}|t|$, and let $\kappa\in \N_+$. Let $(w_*,b_*)\in\R^2$ and,
    for all $l\in\{1, \dots, \kappa\}$, let $(w_l,b_l)\in\R^2$, $\va_l\in\R^{d'}$ and
    $u_l(t)=w_lt+b_l$. Define
    \[
        N_q(t)=\sum_{l=1}^{\kappa}\va_l\prod_{r=1}^q(u_l(t)-u_r(t)),
        \quad q\in\{0, \dots, \kappa-1\},
        \qquad
        \delta=\max_{l\in\intran{1}{\kappa}}\bigl(R|w_l-w_*|+|b_l-b_*|\bigr).
    \]
    Then $N_q\in\bbP_q(J,\R^{d'})$ for each $q$, and
    \begin{equation}
        \label{eq:properness_cluster_lemma}
        \Bigl\|\sum_{l=1}^{\kappa}\va_l\tanh(u_l)-\sum_{q=0}^{\kappa-1}
        \frac{N_q}{q!}\tanh^{(q)}(w_*\cdot+b_*)\Bigr\|_{\Linf(J)}
        \le C_{\kappa,J}\,\delta\max_{q\in\intran{0}{\kappa-1}}\|N_q\|_\coef,
    \end{equation}
    where $\| \cdot \|_\coef$ is defined in \eqref{eq:hashnorm}%
    , and
    \begin{equation}
        \label{eq:CkJ}
        C_{\kappa,J}=\sum_{q=0}^{\kappa-1}
        \frac{\|\tanh^{(q+1)}\|_{\Linf(\R)}}{q!}\sum_{a=0}^qR^a.
    \end{equation}
\end{lemma}
\begin{proof}
    Each $N_q$ is an $\R^{d'}$-valued polynomial of degree at most $q$, being
    a sum of products of $q$ affine functions of $t$.

    Fix $t\in J$. \Cref{lem:properness_newton}, applied with $h=\tanh$,
    $x_l=u_l(t)$ and $\vc_l=\va_l$, gives
    \[
        \sum_{l=1}^{\kappa}\va_l\tanh(u_l(t))
        \overset{\eqref{eq:hermite_newton_vector}}{=}\sum_{q=0}^{\kappa-1}H_q(t)N_q(t),
        \qquad
        \text{where }H_q(t)=\tanh[u_1(t),\ldots,u_{q+1}(t)].
    \]
    Since $|t|\le R$, the triangle
    inequality gives
    \begin{equation}
        \label{eq:ul-delta}
        |u_l(t)-(w_*t+b_*)|
        \le|w_l-w_*|\,|t|+|b_l-b_*|\le\delta,
        \qquad \text{for all }l\in\intran{1}{\kappa}.
    \end{equation}
    All derivatives of $\tanh$ are bounded on $\R$: starting from
    $\tanh'=1-\tanh^2$, by induction each derivative is a
    polynomial in $\tanh$, and $\|\tanh\|_{\rmL^\infty(\R)}=1$. Thus
    \Cref{lem:properness_newton} with $x^*=w_*t+b_*$ gives
    \[
        \begin{split}
            \left|H_q(t)-\frac{\tanh^{(q)}(w_*t+b_*)}{q!}\right|
             & \overset{\eqref{eq:dd_approx}}{\le}\frac{\|\tanh^{(q+1)}\|_{\Linf(\R)}}{q!}
            \max_{l\le q+1}|u_l(t)-(w_*t+b_*)|                                                   \\
             & \overset{\eqref{eq:ul-delta}}{\le}\frac{\|\tanh^{(q+1)}\|_{\Linf(\R)}}{q!}\delta.
        \end{split}
    \]
    Moreover, writing $N_q(t)=\sum_{a=0}^q\vc_at^a$, we have
    \begin{equation*}
        \|N_q(t)\|_{\ell^\infty}\le\sum_{a=0}^q\|\vc_a\|_{\ell^\infty}R^a
        \le\|N_q\|_\coef\sum_{a=0}^qR^a.
    \end{equation*}
    Therefore, for all $t\in J$,
    \begin{align*}
         & \Bigl\|\sum_{l=1}^{\kappa}\va_l\tanh(u_l(t))-\sum_{q=0}^{\kappa-1}
        \frac{N_q(t)}{q!}\tanh^{(q)}(w_*t+b_*)\Bigr\|_{\ell^\infty}              \\
         & =\Bigl\|\sum_{q=0}^{\kappa-1}N_q(t)
        \Bigl(H_q(t)-\frac{\tanh^{(q)}(w_*t+b_*)}{q!}\Bigr)\Bigr\|_{\ell^\infty} \\
         & \le\delta\max_{0\le q<\kappa}\|N_q\|_\coef
        \sum_{q=0}^{\kappa-1}\frac{\|\tanh^{(q+1)}\|_{\Linf(\R)}}{q!}
        \sum_{a=0}^qR^a.
    \end{align*}
    The right-hand side does not depend on $t\in J$, which proves
    \eqref{eq:properness_cluster_lemma}.
\end{proof}

Before treating the remaining neurons, we define extended
exponential sums.

\begin{definition}
    \label{def:extended_exponential_sum}
    A scalar-valued \emph{extended exponential sum} \cite[Chapter~VI]{braess_nonlinear_1986}
    is a finite sum of functions $Q(t)e^{\lambda t}$, where $Q$ is a
    real polynomial and $\lambda\in\R$, and has a
    unique representation of the form
    \begin{equation*}
        \sum_{i=1}^l Q_i(t)e^{\lambda_i t},
    \end{equation*}
    with distinct exponents ($\lambda_i\neq \lambda_j$ if $i\neq j$) and
    nonzero polynomials $Q_i$. We
    call $\sum_{i=1}^l(\deg Q_i+1)$ its \emph{order}. A vector-valued
    function is an extended exponential sum if each of its coordinates is.
    For an interval $J\subset\R$, we denote the set of $\R^{d'}$-valued
    extended exponential sums on $J$ by
    \begin{equation}
        \label{eq:expsum}
        \bbE\bbP(J, \R^{d'}) \coloneqq \left\{ f = \sum_{i=1}^l Q_i(t)e^{\lambda_i t}: l, q\in \N;\; \forall i,\; \lambda_i\in \R, \, Q_i\in \bbP_{q}(J, \R^{d'})\right\}.
    \end{equation}
\end{definition}
The polynomial factors in the definition above arise naturally
when taking limits of sums of exponentials whose exponents approach
one another. For example,
\[
    \frac{e^{(\lambda+h)t}-e^{\lambda t}}{h}
    \longrightarrow t e^{\lambda t}
    \qquad\text{as }h\to0,
\]
uniformly on bounded intervals. The next lemma shows that extended
exponential sums capture all possible $\rmL^1$ limits of sums of
exponentials with a fixed number of terms.
\begin{lemma}
    \label{lem:properness_exp_limits}
    Let $J$ be a nonempty bounded open interval and $L\in \N_+$.
    Let $( \vc_{l,m} )_{l=1}^L\subset \R^{d'}$ and $(
        \lambda_{l,m})_{l=1}^L\subset \R$
    for all $m\in\N_+$, and let
    \[
        Z_m(t)=\sum_{l=1}^L\vc_{l,m}e^{\lambda_{l,m}t}.
    \]
    If there exists $Z$ such that
    $Z_m\to Z$ in $\Lone(J,\R^{d'})$, then $Z$ agrees almost everywhere on $J$ with an element of
    $\bbE\bbP(J,\R^{d'})$.
\end{lemma}
\begin{proof}
    Write $J=(a,b)$ and define, for $t\in[a,b]$,
    \[
        Y_m(t)=\int_a^t Z_m(s)\,ds,\qquad
        Y(t)=\int_a^t Z(s)\,ds.
    \]
    Then $\|Y_m-Y\|_{\Linf([a,b])}\le\|Z_m-Z\|_{\Lone(J)}\to0$, so the
    sequence $Y_m$ is uniformly bounded on $[a,b]$ and $Y_m$ converges to $Y$
    uniformly on $J$. Each coordinate of $Y_m$
    is an extended exponential sum of order at most $L+1$.
    By \cite[Chapter~VI, Theorem~2.3]{braess_nonlinear_1986}, applied to each
    coordinate, the uniform limit $Y$ of $(Y_m)_m$ coincides with an extended
    exponential sum $P$ on every compact subinterval of $J$, hence on $J$.
    Since $Y$ is absolutely continuous with $Y'=Z$ almost
    everywhere, and the derivative of an extended exponential sum is again an
    extended exponential sum, $Z$ agrees almost everywhere on $J$ with an
    element of $\bbE\bbP(J,\R^{d'})$.
\end{proof}
So far in this section, we have studied clusters of neurons whose weights
and biases approach a common finite pair with nonzero weight. We now turn to the neurons whose
weights tend to zero or whose weights or biases diverge. We call these
neurons \emph{degenerate}.
In the notation of \Cref{lem:properness_elimination} below, each of them converges, uniformly on bounded intervals, to the constant $\tanh(b_j^*)$ if
$w_{j,m}\to0$ and $b_{j,m}\to b_j^*\in\R$, and to $\pm1$ if $w_{j,m}$ is bounded
and $b_{j,m}\to\pm\infty$; if $w_{j,m}\to+\infty$, it converges pointwise, away
from at most one point, to a step function. Their amplitudes, however, may
diverge, and then their sum may converge to a function that is neither constant
nor a step function. For example, as $b\to+\infty$,
\[
    e^{2b}\bigl(\tanh(t+b)-1\bigr)
    =\frac{-2e^{-2t}}{1+e^{-2(t+b)}}\longrightarrow-2e^{-2t}
\]
uniformly on bounded intervals: a neuron and the offset, with amplitudes $\pm
    e^{2b}$, converge to an exponential. The next lemma shows that, on a subinterval
containing no discontinuities of the limit step functions, such limits are always extended exponential sums. The key
step of its proof is to write each $\tanh$ as a rational function of a single
exponential and to multiply by the common denominator, which converges to a
positive constant.

\begin{lemma}
    \label{lem:properness_elimination}
    Let $I\subset\R$ be a nonempty bounded open interval and let $E$ be a
    finite set. For all $j\in E$, let $(w_{j,m})_m\subset\Rnonneg$ and
    $(b_{j,m})_m\subset\R$ converge to $w_j^*\in[0,+\infty]$ and
    $b_j^*\in[-\infty,+\infty]$, with $(w_j^*,b_j^*)\notin(0,\infty)\times\R$.
    Then, there exists a subsequence of
    $(\{w_{j,m},b_{j,m}:j\in E\})_{m}$ (not relabeled to ease notation) and a nonempty
    open interval $J$ with $\overline J\subset I$
    such that the following holds along this subsequence and all its
    subsequences.
    If $\vs_m\in\R^{d'}$ and
    $\va_{j,m}\in\R^{d'}$, $j\in E$, are such that
    \[
        f_m(t)\coloneqq\vs_m+\sum_{j\in E}\va_{j,m}\tanh(w_{j,m}t+b_{j,m})
        \longrightarrow f\qquad\text{in }\Lone(J,\R^{d'}),
    \]
    then $f$ agrees almost everywhere on $J$ with an element of
    $\bbE\bbP(J,\R^{d'})$.
\end{lemma}
\begin{proof}
    \emph{Step 1: choice of $J$ and of the signs.}
    Since $(w_j^*,b_j^*)\notin(0,\infty)\times\R$ for every $j\in E$, the sets
    \[
        E_0=\{j:w_j^*=0,\ b_j^*\in\R\},\qquad
        E_B=\{j:w_j^*<\infty,\ |b_j^*|=\infty\},\qquad
        E_W=\{j:w_j^*=+\infty\}
    \]
    form a partition of $E$.
    Denote
    \[
        u_{j,m}(t)=w_{j,m}t+b_{j,m}.
    \]
    For $j\in E_W$ the weights are eventually positive; we may assume
    $w_{j,m}>0$ for all $m$ and introduce, up to a subsequence,
    \[
        c_{j,m}\coloneqq-\frac{b_{j,m}}{w_{j,m}}\qquad
        c_j^*\coloneqq
        \lim_{m\to\infty} c_{j, m}\in[-\infty,+\infty],
    \]
    so that $u_{j,m}(t)=w_{j,m}(t-c_{j,m})$.
    This is the subsequence of the statement: it depends only on the
    weights and biases, and it is needed because $(c_{j,m})_m$ need not
    converge.
    Now, since for all $j\in E_W$, $\tanh(u_{j,m}(t))$ converges to $\pm1$ for every $t\ne
        c_j^*$, choose a nonempty open interval $J$ with $\overline{J}\subset
        I$ that does not contain any of the $c_j^*$, i.e.,
    \begin{equation*}
        \overline{ J }\cap \{c_j^*: j\in E_W\} = \emptyset.
    \end{equation*}
    Let then $R\coloneqq \sup_{t\in
            J}|t|<\infty$.
    We now fix a sign $\epsilon_j\in\{-1,+1\}$ for every $j\in E$:
    \begin{itemize}
        \item For $j\in E_0$, let $\epsilon_j=-1$. For $j\in E_B$, let $\epsilon_j$
              be the sign of $b_j^*=\pm\infty$, so that $\epsilon_jb_{j,m}=|b_{j,m}|$
              eventually.
        \item For $j\in E_W$ with $c_j^*$ finite, let $\epsilon_j$ be the sign of
              $t-c_j^*$ for $t\in\overline J$, which is constant since
              $c_j^*\notin\overline J$.
        \item For $j\in E_W$ with $c_j^*=\pm\infty$, let $\epsilon_j=\mp1$, so that
              $-\epsilon_jc_{j,m}=|c_{j,m}|$ eventually.
    \end{itemize}
    Then,
    \begin{equation}
        \label{eq:properness_saturation}
        \inf_{t\in\overline J}\epsilon_j u_{j,m}(t)\to+\infty,
        \qquad \text{for all }j\in E_B\cup E_W.
    \end{equation}
    Indeed, for $j\in E_B$, since $|t|\le R$ on $\overline J$ and
    $w_{j,m}\to w_j^*<\infty$, for sufficiently large $m$,
    \[
        \epsilon_ju_{j,m}(t)=\epsilon_jb_{j,m}+\epsilon_jw_{j,m}t
        \ge|b_{j,m}|-Rw_{j,m}\to+\infty.
    \]
    For $j\in E_W$, factoring out the weight,
    \[
        \inf_{t\in\overline J}\epsilon_ju_{j,m}(t)
        =w_{j,m}\inf_{t\in\overline J}\epsilon_j(t-c_{j,m}).
    \]
    If $c_j^*$ is finite, then
    $|\epsilon_j(t-c_{j,m})-\epsilon_j(t-c_j^*)|=|c_{j,m}-c_j^*|\to0$
    uniformly in $t$, so the infimum on the right converges to
    $\inf_{t\in\overline J}|t-c_j^*|=\operatorname{dist}(c_j^*,\overline J)>0$.
    If $c_j^*$ is infinite, then
    $\epsilon_j(t-c_{j,m})\ge|c_{j,m}|-R\to+\infty$. In both cases, the
    infimum on the right is eventually bounded below by a positive
    constant, and $w_{j,m}\to+\infty$ proves
    \eqref{eq:properness_saturation}.

    \emph{Step 2: the common denominator.}
    The set $E$ may be empty; in this case, all the formulas below remain valid
    under the convention that empty sums and products equal $0$ and $1$,
    respectively. Define
    \[
        D_m(t)\coloneqq \prod_{j\in E}(1+e^{-2\epsilon_j u_{j,m}(t)}).
    \]
    For every fixed $j\in E_B\cup E_W$, \eqref{eq:properness_saturation} gives
    \[
        \sup_{t\in\overline J}\left|e^{-2\epsilon_j u_{j,m}(t)}\right|
        \to0,
    \]
    so that $e^{-2\epsilon_j u_{j,m}(t)}\to 0$ uniformly on $\overline J$. In addition, for every fixed $j\in E_0$ we have
    \[
        \sup_{t\in\overline J}|u_{j,m}(t)-b_j^*|
        \le Rw_{j,m}+|b_{j,m}-b_j^*|\to0,
    \]
    so that, since $\epsilon_j=-1$ for $j\in E_0$,
    $e^{-2\epsilon_j u_{j,m}}=e^{2u_{j,m}}\to e^{2b_j^*}$
    uniformly on $\overline J$.
    Therefore, given that $E$ is finite, the product satisfies
    \begin{equation}
        \label{eq:properness_denominator}
        D_m\to D_0\coloneqq\prod_{j\in E_0}(1+e^{2b_j^*})>0
        \quad\text{uniformly on }\overline J.
    \end{equation}
    In particular $D_m\ge1$, the sequence $(D_m)_m$ is uniformly bounded on
    $J$, and $D_0$ is a constant.
    \emph{Step 3: exact form.}
    Let $\vs_m$, $\va_{j,m}$, and $f_m$ be as in the statement, and set
    \[
        Z_m\colon\overline J\to\R^{d'},\qquad
        Z_m(t)\coloneqq D_m(t)f_m(t).
    \]
    We now expand $Z_m$ as a sum of exponentials. Indeed, for either sign
    $\epsilon = \pm 1$, the identity
    \[
        \tanh u=\epsilon\frac{1-e^{-2\epsilon u}}{1+e^{-2\epsilon u}}
    \]
    gives
    \begin{align*}
        Z_m(t)
         & = \Big(\prod_{l\in E}(1+e^{-2\epsilon_l u_{l,m}(t)})\Big)
        \cdot \Big(\vs_m+\sum_{j\in E}\va_{j,m}\tanh(u_{j,m}(t))\Big)       \\
         & = \vs_m\prod_{l\in E}(1+e^{-2\epsilon_l u_{l,m}(t)})
        +\Big(\prod_{l\in E}(1+e^{-2\epsilon_l u_{l,m}(t)})\Big)
        \sum_{j\in E}\epsilon_j\va_{j,m}
        \frac{1-e^{-2\epsilon_j u_{j,m}(t)}}{1+e^{-2\epsilon_j u_{j,m}(t)}} \\
         & = \vs_m\prod_{l\in E}(1+e^{-2\epsilon_l u_{l,m}(t)})
        +\sum_{j\in E}\epsilon_j\va_{j,m}(1-e^{-2\epsilon_j u_{j,m}(t)})
        \prod_{l\in E\setminus\{j\}}(1+e^{-2\epsilon_l u_{l,m}(t)}).
    \end{align*}
    We expand the products using the following identities,
    valid for any real numbers $(x_l)_{l\in E}\subset\R$ and $j\in E$,
    \[
        \prod_{l\in E}(1+x_l)=\sum_{S\subseteq E}\prod_{l\in S}x_l,
        \qquad
        (1-x_j)\prod_{l\in E\setminus\{j\}}(1+x_l)
        =\sum_{\substack{S\subseteq E\\j\notin S}}\prod_{l\in S}x_l
        -\sum_{\substack{S\subseteq E\\j\in S}}\prod_{l\in S}x_l,
    \]
    where the second identity follows from the first, applied to
    $E\setminus\{j\}$. With
    \[
        \lambda_{S,m}\coloneqq -2\sum_{j\in S}\epsilon_jw_{j,m},
        \qquad
        \nu_{S,m}\coloneqq -2\sum_{j\in S}\epsilon_jb_{j,m},
        \qquad \text{for }S\subseteq E,
    \]
    we have
    $\prod_{j\in S}e^{-2\epsilon_j u_{j,m}(t)}
        =e^{\nu_{S,m}}e^{\lambda_{S,m}t}$,
    and therefore
    \begin{align*}
        Z_m(t)
         & = \sum_{S\subseteq E}e^{\nu_{S,m}}e^{\lambda_{S,m}t}\vs_m
        +\sum_{j\in E}\epsilon_j\va_{j,m}\Biggl(
        \sum_{\substack{S\subseteq E                                 \\j\notin S}}
        e^{\nu_{S,m}}e^{\lambda_{S,m}t}
        -\sum_{\substack{S\subseteq E                                \\j\in S}}
        e^{\nu_{S,m}}e^{\lambda_{S,m}t}\Biggr)                       \\
         & = \sum_{S\subseteq E}e^{\nu_{S,m}}
        \Biggl(\vs_m+\sum_{j\in E\setminus S}\epsilon_j\va_{j,m}
        -\sum_{j\in S}\epsilon_j\va_{j,m}\Biggr)e^{\lambda_{S,m}t}
        =\sum_{S\subseteq E}\veta_{S,m}e^{\lambda_{S,m}t},
    \end{align*}
    where
    \begin{equation}
        \label{eq:properness_exact_rates}
        \veta_{S,m}=
        e^{-2\sum_{j\in S}\epsilon_jb_{j,m}}
        \Biggl(\vs_m+\sum_{j\in E}\epsilon_j\va_{j,m}
        -2\sum_{j\in S}\epsilon_j\va_{j,m}\Biggr).
    \end{equation}
    Thus $Z_m$ is a sum of at most $2^{|E|}$ real exponentials with
    coefficients in $\R^{d'}$; some rates $\lambda_{S,m}$ may coincide and
    some coefficients may vanish, which only reduces the number of terms.

    \emph{Step 4: passing to the limit.} Suppose that $f_m\to f$ in $\Lone(J,\R^{d'})$ along the subsequence
    of Step 1 or any further subsequence, still indexed by $m$.
    By Step 3, each $Z_m=D_mf_m$ is a sum of at most $2^{|E|}$ real
    exponentials with coefficients in $\R^{d'}$. Moreover,
    \[
        \|Z_m-D_0f\|_{\Lone(J)}
        \le\|D_m\|_{\Linf(J)}\|f_m-f\|_{\Lone(J)}
        +\|D_m-D_0\|_{\Linf(J)}\|f\|_{\Lone(J)}\longrightarrow0,
    \]
    since $(D_m)_m$ is uniformly bounded and $D_m\to D_0$ uniformly on
    $\overline J$ by \eqref{eq:properness_denominator}. By
    \Cref{lem:properness_exp_limits} with $L=2^{|E|}$, $D_0f$ agrees almost
    everywhere on $J$ with an element of $\bbE\bbP(J,\R^{d'})$, and so does $f$,
    since $D_0>0$ is a constant.
\end{proof}

The next proposition describes, up to a subsequence, the $\Lone$ limits of
one-dimensional networks with $n$ neurons on a suitable subinterval.
In particular, if a sequence of networks converges in $\Lone$, its limit is
the sum of (i) an extended exponential sum and (ii) the products of
polynomials and derivatives of $\tanh$. The maximum polynomial degree and the
order of differentiation of the latter terms
depend on the multiplicity of the limit points of
the sequence of weights and biases.

\begin{proposition}[Local structure of one-dimensional limits]
    \label{prop:properness_local_structure}
    Let $I\subset\R$ be a nonempty bounded open interval. Let
    $\vs_m\in \R^{d'}$, $(\va_{j,m})_{j=1}^n\subset \R^{d'}$,
    $(w_{j,m})_{j=1}^n\subset\Rnonneg$, and $(b_{j,m})_{j=1}^n\subset \R$ for all $m\in\Npos$.
    Suppose that there exists $g$ such that
    \[
        g_m(t)=\vs_m+\sum_{j=1}^n\va_{j,m}\tanh(w_{j,m}t+b_{j,m})\to g \qquad
        \text{in }\Lone(I, \R^{d'}).
    \]
    Then,
    there exist
    a subsequence $(m_k)_k$,
    a family $\calK$ of pairwise
    disjoint nonempty subsets of $\{1,\ldots,n\}$, called clusters, with
    pairwise distinct centers $(w_K,b_K)\in(0,\infty)\times\R$, $K\in\calK$,
    a nonempty open interval $J$ with $\overline J\subset I$, an extended
    exponential sum $F\in\bbE\bbP(J,\R^{d'})$, and polynomials $N^K_q\in \mathbb{P}_q(J,
        \R^{d'})$,
    for $K\in\calK$ and $q\in\intran{0}{|K|-1}$, such that:
    \begin{enumerate}[(i)]
        \item for every $K\in\calK$ and $j\in K$,
              $(w_{j,m_k},b_{j,m_k})\to(w_K,b_K)$; moreover,
              $\sum_{j\in K}\va_{j,m_k}\to N^K_0$;
        \item for almost every $t\in J$,
              \begin{equation}
                  \label{eq:properness_local_structure}
                  g(t)=F(t)+\sum_{K\in\calK}\sum_{q=0}^{|K|-1}
                  \frac{N^K_q(t)}{q!}\tanh^{(q)}(w_Kt+b_K).
              \end{equation}
    \end{enumerate}
\end{proposition}
\begin{proof}
    Since we extract subsequences several times below, with a slight abuse of
    notation we keep the index $m$ for every subsequence.

    \emph{Step 1: splitting of the neurons.}
    Up to a subsequence, we may assume that for all $j\in\intran{1}{n}$,
    the parameters $(w_{j, m})_m$ and $(b_{j,m})_m$
    have a (finite or infinite) limit, and denote it as follows:
    \[
        w_{j,m}\to w_j^*\in[0,+\infty],\qquad
        b_{j,m}\to b_j^*\in[-\infty,+\infty].
    \]
    All subsequent extractions preserve these limits. Let
    \[
        I_C^+=\{j:0<w_j^*<\infty,\ b_j^*\in\R\},\qquad
        E=\{1,\ldots,n\}\setminus I_C^+,
    \]
    and $u_{j,m}(t)=w_{j,m}t+b_{j,m}$. The neurons in $E$ satisfy the
    hypotheses of \Cref{lem:properness_elimination}, which provides, 
    a subsequence, a nonempty
    open interval $J$ with $\overline J\subset I$, depending only on the weights
    and biases of these neurons and valid along the further subsequences
    extracted below. Set
    $R=\sup_{t\in J}|t|<\infty$ and
    \[
        G_m(t)=\sum_{j\in I_C^+}\va_{j,m}\tanh(u_{j,m}(t)),
    \]
    so that $g_m-G_m=\vs_m+\sum_{j\in E}\va_{j,m}\tanh(u_{j,m})$ has the form
    considered in \Cref{lem:properness_elimination}.

    We suppose temporarily that $I_C^+$ is not empty. Group $I_C^+$ into clusters $K$ of indices whose pairs
    $(w_{i,m},b_{i,m})$ have a common limit $(w_K,b_K)\in(0,\infty)\times\R$,
    and let $\calK$ be the family of these clusters. Different clusters have
    different centers $(w_K,b_K)$.
    For each cluster $K=\{i_1,\ldots,i_{|K|}\}$, let
    \[
        N^K_{q,m}(t)=\sum_{l=1}^{|K|}\va_{i_l,m}
        \prod_{r=1}^q(u_{i_l,m}(t)-u_{i_r,m}(t)),
        \qquad 0\le q<|K|.
    \]
    By \Cref{lem:properness_cluster}, each $N^K_{q,m}\in\bbP_q(J,\R^{d'})$, and
    \begin{equation}
        \label{eq:properness_cluster_amplitude}
        N^K_{0,m}=\sum_{i\in K}\va_{i,m}.
    \end{equation}
    Set
    \[
        \begin{split}
            \delta_m & =\max_{K\in \calK}\max_{j\in K}
            \bigl(R|w_{j,m}-w_K|+|b_{j,m}-b_K|\bigr)\to0,
            \\
            \mu_m    & =\max_{K\in \calK}\max_{q\in\intran{0}{\left|K\right|-1}}\|N^K_{q,m}\|_\coef.
        \end{split}
    \]
    \Cref{lem:properness_cluster}, applied to each cluster $K$ with center
    $(w_*,b_*)=(w_K,b_K)$, gives the bound
    \begin{equation*}
        \Bigl\|\sum_{j\in K}\va_{j,m}\tanh(u_{j,m}(\cdot))-\sum_{q=0}^{|K|-1}
        \frac{N^K_{q,m}}{q!}\tanh^{(q)}(w_K\cdot+b_K)\Bigr\|_{\Linf(J)}
        \le
        C_{|K|,J}\delta_m\mu_m,
    \end{equation*}
    with $C_{|K|,J}$ defined as in \eqref{eq:CkJ}.
    There are at most $n$ clusters, and
    $C_{|K|,J}\le C_{n,J}$ since $|K|\le n$; summing over the clusters,
    \begin{equation}
        \label{eq:properness_cluster_error}
        \Bigl\|G_m-\sum_K\sum_{q=0}^{|K|-1}
        \frac{N^K_{q,m}}{q!}\tanh^{(q)}(w_K\cdot+b_K)
        \Bigr\|_{\Linf(J)}
        \le nC_{n,J}\delta_m\mu_m.
    \end{equation}
    If $I_C^+$ is empty and
    there are no clusters, $G_m=0$ and we set $\mu_m=\delta_m=0$.

    \emph{Step 2: boundedness of $\mu_m$ (and hence of the polynomial coefficients).}
    We show this by contradiction.
    Suppose $( \mu_m )_m$
    is unbounded, and extract a subsequence so that $\mu_m\to\infty$.
    Each $N^K_{q,m}\in\bbP_q(J,\R^{d'})$, where $q<|K|\le n$ does not depend on $m$, and
    $\|N^K_{q,m}/\mu_m\|_\coef\le1$ by definition of $\mu_m$.
    It follows that for all $K\in\calK$ and all $q\in \intran{0}{|K|-1}$,
    there exist polynomials $\widehat{N}^K_q$ of degree at most
    $q$ such that,
    after a further extraction,
    \[
        N^K_{q,m}/\mu_m\to\widehat N^K_q
        \quad\text{coefficientwise},\qquad
        \max_{K,q}\|\widehat N^K_q\|_\coef=1.
    \]
    Since $J$ is bounded, coefficientwise convergence implies convergence in
    $\Linf(J)$. Dividing
    \eqref{eq:properness_cluster_error} by $\mu_m$ and using $\delta_m\to0$, we
    obtain
    \begin{equation}
        \label{eq:properness_Ghat}
        \frac{G_m}{\mu_m}\to\widehat G\coloneqq
        \sum_K\sum_{q=0}^{|K|-1}
        \frac{\widehat N^K_q}{q!}\tanh^{(q)}(w_K\cdot+b_K)
        \quad\text{in }\Linf(J,\R^{d'}).
    \end{equation}
    In addition, $g_m$ converges in $\Lone(I)$, so
    $\|g_m\|_{\Lone(J)}$ is bounded and $g_m/\mu_m\to0$ in $\Lone(J)$.
    It follows that
    \begin{equation*}
        \frac{g_m-G_m}{\mu_m}\to-\widehat G \qquad \text{in }\Lone(J).
    \end{equation*}
    The functions $(g_m-G_m)/\mu_m$ have the form considered in
    \Cref{lem:properness_elimination}, with offset $\vs_m/\mu_m$ and amplitudes
    $\va_{j,m}/\mu_m$, $j\in E$. Hence $-\widehat G$, and thus $\widehat G$,
    agrees almost everywhere on $J$ with an element of $\bbE\bbP(J,\R^{d'})$.
    It can therefore be written as
    \begin{equation}
        \label{eq:properness_Ghat_2}
        \widehat{G}(t) = Q_0(t)+\sum_{v=1}^hQ_v(t)e^{\lambda_vt}
    \end{equation}
    with distinct
    nonzero rates $\lambda_v$ and vector-valued polynomials $Q_v$.
    Comparing \eqref{eq:properness_Ghat} and \eqref{eq:properness_Ghat_2},
    since
    both sides
    are continuous,
    \begin{equation}
        \label{eq:properness_Ghat_exp}
        \sum_K\sum_{q=0}^{|K|-1}
        \frac{\widehat N^K_q(t)}{q!}\tanh^{(q)}(w_Kt+b_K)
        -Q_0(t)-\sum_{v=1}^hQ_v(t)e^{\lambda_vt}=\mathbf0
        \qquad \forall t\in J.
    \end{equation}
    Since the pairs $(w_K, b_K)$ are distinct, and so are the exponents
    $\lambda_v$, by \Cref{lem:generalized_independence}
    all the
    polynomials in \eqref{eq:properness_Ghat_exp} must vanish; in particular
    $\widehat N^K_q=0$ for all $K$ and $q$. This contradicts
    $\max_{K,q}\|\widehat N^K_q\|_\coef=1$, and therefore $\mu_m$ is
    bounded.

    \emph{Step 3: passing to the limit.}
    Since $\mu_m$ is bounded, the coefficients of the polynomials
    $N^K_{q,m}$ are bounded, and we may extract coefficientwise limits
    $N^K_{q,m}\to N^K_q$; each $N^K_q\in\bbP_q(J,\R^{d'})$. As in Step 2,
    \eqref{eq:properness_cluster_error} and $\delta_m\to0$ imply
    \[
        G_m\to G_\infty\coloneqq\sum_K\sum_{q=0}^{|K|-1}
        \frac{N^K_q}{q!}\tanh^{(q)}(w_K\cdot+b_K)
        \quad\text{in }\Linf(J,\R^{d'}).
    \]
    Since in addition $g_m\to g$ in $\Lone(J)$, we have $g_m-G_m\to
        g-G_\infty$ in $\Lone(J)$, and \Cref{lem:properness_elimination} shows
    that
    $g-G_\infty$  agrees almost everywhere on
    $J$ with an element $F$ of $\bbE\bbP(J,\R^{d'})$. This proves (ii). Finally,
    (i) follows from the definition of the clusters and from
    \eqref{eq:properness_cluster_amplitude}, since $N^K_{0,m}\to N^K_0$.
\end{proof}

We now have all the ingredients to prove the one-dimensional case.

\begin{lemma}[One-dimensional inverse stability]
    \label{lem:properness_interval}
    Let $I\subset\R$ be a nonempty bounded open interval and $1\le p\le\infty$.
    Let $\vs_m\in \R^{d'}$, $(\va_{j,m})_{j=1}^n\subset \R^{d'}$,
    $(w_{j,m})_{j=1}^n\subset\R$, and $(b_{j,m})_{j=1}^n\subset \R$ for all $m\in\Npos$.
        Suppose that there exists an irreducible NN $g$ of hidden dimension $n$ such that
    \begin{equation}
        \label{eq:properness_interval_hyp}
        g_m\coloneqq\vs_m+\sum_{j=1}^n\va_{j,m}\tanh(w_{j,m}\cdot +b_{j,m})
        \longrightarrow
        g
    \end{equation}
    in $\Lp(I,\R^{d'})$.
    Then the parameter sequence
    \begin{equation*}
        \{\vtheta_m\defas\bigl((w_{j,m},b_{j,m},\va_{j,m})_{j=1}^n,\vs_m\bigr)\}_{m\in\N}
    \end{equation*}
    is bounded.
\end{lemma}
\begin{proof}
    We write
    \begin{equation}
        \label{eq:properness_interval_g1}
        g(t)=\vs+\sum_{i=1}^n\va_i\tanh(w_it+b_i),
    \end{equation}
    where $\vs\in\R^{d'}$, $w_i\ne0$,
    $\va_i\ne\mathbf0$, and $b_i\in \R$, for $i\in\intran{1}{n}$, and
    the pairs $(w_i,b_i)$ are distinct
    up to sign (by the hypothesis of irreducibility). We prove the lemma by contradiction.
    Suppose that $\{\vtheta_m\}$ is unbounded. Then, there exists a subsequence
    with $\|\vtheta_{m_k}\|\to\infty$ as $k\to\infty$.
    To reach a contradiction, it is therefore sufficient to find a further
    subsequence of $\{\vtheta_{m_k}\}$ that is bounded.
    Since we extract
    subsequences several times below, with a slight abuse of notation we keep
    the name $\vtheta_m$, and likewise $g_m$, $w_{j,m}$, $b_{j,m}$, $\va_{j,m}$
    and $\vs_m$, for every subsequence.

    Since $\tanh$ is odd, changing the signs of an entire triple
    $(w_{j,m},b_{j,m},\va_{j,m})$ changes neither $g_m$ nor the parameter norm.
    We may therefore assume $w_{j,m}\ge0$ for all $j$ and $m$ and, in the same
    way, $w_i>0$ for all $i$. Finally, since $I$ is bounded,  $g_m\to g$ also in $\Lone(I,\R^{d'})$.

    We may therefore apply \Cref{prop:properness_local_structure}.
    There exist a family $\calK$ of
    disjoint nonempty subsets of $\{1,\ldots,n\}$,
    distinct $(w_K,b_K)\in(0,\infty)\times\R$,
    an open interval $J$ with $\overline J\subset I$, an extended
    exponential sum $F$, and polynomials $N^K_q\in \mathbb{P}_q(J,
        \R^{d'})$ such that, passing to a subsequence, $( w_{j,m}, b_{j,m} )\to
        (w_K, b_K)$ if $j\in K$ and
    \begin{equation}
        \label{eq:properness_interval_g2}
        g_m\to F(\cdot)+\sum_{K\in\calK}\sum_{q=0}^{|K|-1}
        \frac{N^K_q(\cdot)}{q!}\tanh^{(q)}(w_K\cdot+b_K)\qquad \text{in }\Lone(J).
    \end{equation}

    We write
    $F(t)=Q_0(t)+\sum_{v=1}^hQ_v(t)e^{\lambda_vt}$ with distinct nonzero
    rates $\lambda_v$ and polynomials $Q_v$.
    Comparing \eqref{eq:properness_interval_hyp},
    \eqref{eq:properness_interval_g1}, and \eqref{eq:properness_interval_g2},
    we infer that, for all $t\in J$ by continuity,
    \begin{equation}
        \label{eq:properness_matching_identity}
        \begin{split}
             & \vs-Q_0(t)+\sum_{i=1}^n\va_i\tanh(w_it+b_i)
            -\sum_{K\in\calK}\sum_{q=0}^{|K|-1}
            \frac{N^K_q(t)}{q!}\tanh^{(q)}(w_Kt+b_K)              \\
             & \qquad -\sum_{v=1}^hQ_v(t)e^{\lambda_vt}=\mathbf0.
        \end{split}
    \end{equation}
    All the pairs $(w_i,b_i)$ and $(w_K,b_K)$ have positive first
    entries.
    We now wish to show that \eqref{eq:properness_matching_identity} requires
    \begin{equation}
        \label{eq:omegai_in_omegaK}
        \{( w_i, b_i ) : i\in\intran{1}{n}\} \subset \{(w_K, b_K): K\in \calK\} \eqqcolon C_{\calK}.
    \end{equation}
    Suppose by contradiction that for some $i_0\in\intran{1}{n}$,
    $(w_{i_0},b_{i_0})\not\in C_{\calK}$.
    Since $\va_{i_0}\neq\mathbf0$, the
    linear independence of \Cref{lem:generalized_independence} gives a
    contradiction. Hence \eqref{eq:omegai_in_omegaK} holds.

    Since the $\{(w_i, b_i)\}_{i=1}^n$ are distinct,
    \[
        n\le\#\calK\le\sum_{K\in\calK}|K|\le n,
    \]
    where the second inequality holds because $|K|\ge1$, and the third
    because the clusters are pairwise disjoint subsets of
    $\{1,\ldots,n\}$.
    Therefore,
    \begin{equation*}
        \{( w_i, b_i ) : i\in\intran{1}{n}\} =  C_{\calK} \quad \text{and}\quad \calK = \{\{1\}, \dots, \{n\}\}.
    \end{equation*}%
    By \Cref{prop:properness_local_structure}(i),
    $(w_{j,m},b_{j,m})\to(w_K,b_K)$ and $\va_{j,m}\to N^{\{j\}}_0$; in particular,
    all weights, biases and amplitudes are bounded uniformly in $m$. Finally, for $t\in J$,
    \begin{equation}
        \label{eq:sm=stuff}
        \vs_m=g_m(t)-\sum_{j=1}^n\va_{j,m}\tanh(w_{j,m}t+b_{j,m}).
    \end{equation}
    Since $\|\tanh\|_{\rmL^\infty(\R)}=1$, integrating over $J$ the $\ell^1$ norm
    of the left- and right-hand sides of \eqref{eq:sm=stuff} gives
    \[
        |J|\,\|\vs_m\|_{\ell^1}=\|\vs_m\|_{\Lone(J)}
        \le\|g_m\|_{\Lone(J)}+|J|\sum_{j=1}^n\|\va_{j,m}\|_{\ell^1},
    \]
    which bounds $\vs_m$ uniformly in $m$
    because $g_m$ converges in $\Lone(J)$ and $|J|>0$.
    Thus, a subsequence of the sequence $\{\vtheta_{m_k}\}$ introduced at
    the beginning of the proof is bounded, contradicting the
    initial choice of norms tending to infinity.
\end{proof}

\subsection{From one to several input dimensions}
\label{sec:properness_multid}
The previous lemma proves the inverse stability of the realization map
in the one-dimensional-input case.
To extend this result to several input dimensions,
we first choose a basis $\vv_1,\ldots,\vv_d$ of $\R^d$ such that the
limit network, restricted to any line parallel to some $\vv_k$, is still an
irreducible network with $n$ neurons (\Cref{lem:generic_directions}). We
then reason coordinatewise
in this basis: applying
\Cref{lem:properness_interval} on suitable line slices bounds the
components $\vw_{j,m}^\top\vv_k$ of the weights for every $k$, which allows us
to pass from the one-dimensional case to the general one.

\begin{lemma}
    \label{lem:generic_directions}
    Let $(\vw_1,b_1),\ldots,(\vw_n,b_n)\in\R^d\times\R$ be such that
    \begin{enumerate}[(i)]
        \item $\vw_i\neq\mathbf 0$ for all $i=1,\ldots,n$;
        \item $(\vw_i,b_i)\neq\pm(\vw_j,b_j)$ for all $j\neq i$.
    \end{enumerate}
    Then, there exists a basis $\vv_1,\ldots,\vv_d$ of $\R^d$ such that for
    \emph{every} $\vy\in\R^d$ and every $k\in\{1,\ldots,d\}$ it holds that
    \begin{equation}
        \label{eq:generic_directions}
        \vw_i^\top\vv_k\neq0 \quad \forall i,
        \qquad
        \bigl(\vw_i^\top\vv_k,\ \vw_i^\top\vy+b_i\bigr)
        \neq\pm\bigl(\vw_j^\top\vv_k,\ \vw_j^\top\vy+b_j\bigr)
        \quad \forall i\neq j.
    \end{equation}
\end{lemma}
\begin{proof}
    Define the finite set of nonzero vectors
    \[
        \begin{split}
            \calZ = \{\vw_i\}_{i=1}^{n}
             & \cup\{\vw_i-\vw_j \ : \ \vw_i\neq\vw_j,\ 1\le i,j\le n\}  \\
             & \cup\{\vw_i+\vw_j \ : \ \vw_i\neq-\vw_j,\ 1\le i,j\le n\}
            \subseteq\R^d\setminus\{\mathbf 0\},
        \end{split}
    \]
    which is nonempty and contains no zero vector by hypothesis~(i) and by the
    two conditions imposed on the differences and the sums.

    For every $\vz\in\calZ$ the polynomial $p_{\vz}(t)=\sum_{a=1}^{d}z_at^{a-1}$
    is not identically zero, hence its set of real roots $R_{\vz}\subset\R$
    satisfies $\#R_{\vz}\le d-1$. Since $\calZ$ is finite, $R_{\calZ}
        =\bigcup_{\vz\in\calZ}R_{\vz}$ is finite as well, and we may pick
    pairwise distinct $t_1,\ldots,t_d\in\R\setminus R_{\calZ}$.
    Define the columns of the Vandermonde matrix
    \[
        \vv_k=\bigl(1,\ t_k,\ t_k^2,\ \ldots,\ t_k^{d-1}\bigr)^\top\in\R^d,
        \qquad k=1,\ldots,d.
    \]
    These vectors have the two required properties:
    \begin{enumerate}[(a)]
        \item \label{it:generic_nonzero} $\vv_k^\top\vz=p_{\vz}(t_k)\neq0$ for
              every $\vz\in\calZ$ and every $k$, by the choice of $t_k$;
        \item $\vv_1,\ldots,\vv_d$ is a basis of $\R^d$, because
              $\mV=[\vv_1|\cdots|\vv_d]$ is a Vandermonde matrix with pairwise
              distinct nodes $t_1,\ldots,t_d$, hence invertible.
    \end{enumerate}
    In particular $\vw_i^\top\vv_k\neq0$ for all $i$ and $k$, since
    $\vw_i\in\calZ$, which is the first assertion in
    \eqref{eq:generic_directions}.

    For the second assertion in \eqref{eq:generic_directions}, suppose
    by contradiction that for some $\vy\in\R^d$, some
    $k\in\{1,\ldots,d\}$, some $i\neq j$ and some sign $\varepsilon\in\{-1,+1\}$
    it holds that
    \[
        \bigl(\vw_i^\top\vv_k,\ \vw_i^\top\vy+b_i\bigr)
        =\varepsilon\bigl(\vw_j^\top\vv_k,\ \vw_j^\top\vy+b_j\bigr).
    \]
    Equality of the first term in the tuple requires $\vv_k^\top(\vw_i-\varepsilon\vw_j)
        =0$, so $\vw_i-\varepsilon\vw_j\notin\calZ$ by \eqref{it:generic_nonzero}, and by the definition of
    $\calZ$ this forces $\vw_i=\varepsilon\vw_j$. Equality of the second
    term then gives $\vw_i^\top\vy+b_i=\varepsilon(\vw_j^\top\vy+b_j)
        =\vw_i^\top\vy+\varepsilon b_j$, whence $b_i=\varepsilon b_j$. Altogether
    $(\vw_i,b_i)=\varepsilon(\vw_j,b_j)$, contradicting hypothesis~(ii). The
    conclusion follows since $\vy$ and $k$
    were arbitrary.
\end{proof}

\begin{remark}
    \label{rem:generic_directions_scope}
    If $\vtheta = \{(\vw_i)_i, (b_i)_i, (\va_i)_i, \vs\}$
    is irreducible and
    $\vv_1,\ldots,\vv_d$ are the directions given by the lemma, then for every
    $\vy\in\R^d$ and every $k$ the restricted network
    \begin{align*}
        t\mapsto u_{\vtheta}(\vy+t\vv_k) & =\vs+\sum_{i=1}^{n}\tanh\bigl(\vw_i^\top(\vy+t\vv_k)+b_i\bigr)\va_i \\
                                         & =\vs+\sum_{i=1}^{n}\tanh(\alpha_it+\beta_i)\va_i,
    \end{align*}
    where $\alpha_i\coloneqq\vw_i^\top\vv_k$ and
    $\beta_i\coloneqq\vw_i^\top\vy+b_i$, is an irreducible shallow tanh neural
    network with hidden dimension $n$ in the sense of
    \Cref{def:nondegenerate_tanh_nn}, with $d=1$.
\end{remark}

We are now ready to prove \Cref{thm:boundedness}, which we recall for convenience.
\boundedness*

\begin{proof}
    Since $\Omega$ is bounded, H\"older's inequality gives
    $\|f\|_{\Lone(\Omega)}\le|\Omega|^{1-1/p}\|f\|_{\Lp(\Omega)}$, so
    $u_{\vtheta_m}\to u_{\vtheta}$ also in $\Lone(\Omega,\R^{d'})$. We may
    therefore assume $p=1$, and we prove that $\{\vtheta_m\}$ is bounded
    in $\R^N$ in three steps.

    \textit{Step 1: the case $d=1$.}
    If $d=1$, then $\Omega\subset\R$ is a nonempty open set, hence it contains a
    nonempty bounded open interval $I$ with $\overline I\subset\Omega$.
    Convergence in $\Lone(\Omega,\R^{d'})$ implies convergence in
    $\Lone(I,\R^{d'})$, and the target parameters satisfy $w_i\neq0$,
    $\va_i\neq\mathbf 0$ and $(w_i,b_i)\neq\pm(w_j,b_j)$ for $i\neq j$ by
    irreducibility. These are exactly the hypotheses of
    \Cref{lem:properness_interval}, which therefore yields the boundedness of
    $\{\vtheta_m\}$ directly. In Steps 2 and 3 we therefore assume
    $d\ge2$, and reduce this case to the one-dimensional one by restricting
    the networks to suitably chosen line slices.

    \textit{Step 2: reduction to a line slice.}
    Suppose, by contradiction, that $\{\vtheta_m\}_{m\in\N}$ is unbounded, and
    extract a subsequence $\{\vtheta_{m_q}\}_{q\in\N}$ with
    $\|\vtheta_{m_q}\|\to+\infty$ as $q\to\infty$. Any further subsequence of
    $\{\vtheta_{m_q}\}_{q\in\N}$ still has parameter norms tending to
    $+\infty$; therefore, to reach a contradiction, it suffices to exhibit a
    \emph{bounded} further subsequence. Since $u_{\vtheta_{m_q}}\to u_{\vtheta}$
    in $\Lone(\Omega,\R^{d'})$, we may extract once more so that
    \begin{equation}
        \label{eq:properness_summable_error}
        \|u_{\vtheta_m}-u_{\vtheta}\|_{\Lone(\Omega,\R^{d'})}
        \le2^{-m}\qquad(m\ge1),
    \end{equation}
    where, to ease the notation, we write $\vtheta_m$ in place of the
    twice-extracted subsequence. All further reasoning concerns this
    subsequence.

    The target $\vtheta\in\mTheta$ is irreducible, so its pairs
    $(\vw_i,b_i)$, $i=1,\ldots,n$, satisfy the hypotheses of
    \Cref{lem:generic_directions}. Let then $\vv_1,\ldots,\vv_d$ be a Vandermonde
    basis of $\R^d$ that satisfies \eqref{eq:generic_directions}, and let
    \begin{equation}
        \label{eq:properness_basis_matrix}
        \mV=[\vv_1|\cdots|\vv_d]\in\R^{d\times d},
    \end{equation}
    which is invertible.

    Fix $k\in\{1,\ldots,d\}$ and let $H_k=\vv_k^\perp$ be the hyperplane
    orthogonal to $\vv_k$. For $\vy\in H_k$ define the slice of $\Omega$ along
    the line through $\vy$ with direction $\vv_k$, and the corresponding
    error along this line as
    \begin{equation}
        \label{eq:properness_slice}
        \Omega_{\vy,k}=\{t\in\R:\vy+t\vv_k\in\Omega\},\qquad
        \Upsilon_{m,k}(\vy)=\int_{\Omega_{\vy,k}}
        \|u_{\vtheta_m}(\vy+t\vv_k)-u_{\vtheta}(\vy+t\vv_k)\|_{\ell^1}\,dt.
    \end{equation}
    Note that the
    map $\phi_{\vy,k}(t)=\vy+t\vv_k$ is continuous and $\Omega_{\vy,k}=\phi_{\vy,k}^{-1}(\Omega)$ is open, hence
    measurable, being the preimage of the open set $\Omega$ under $\phi_{\vy,k}$; moreover the integrand in
    \eqref{eq:properness_slice} is continuous and nonnegative. Therefore
    $\Upsilon_{m,k}(\vy)\in[0,+\infty]$ is well defined for every $\vy\in H_k$.

    Define $g_m:\R^d\to\R$ by
    \begin{equation}
        \label{eq:properness_g}
        g_m(\vx)=\|u_{\vtheta_m}(\vx)-u_{\vtheta}(\vx)\|_{\ell^1}\,\mathbf 1_{\Omega}(\vx)
        \ \ge 0.
    \end{equation}
    Let $\{\vq_{1,k},\ldots,\vq_{d-1,k}\}$ be an orthonormal basis of $H_k$ and
    put $\mM=[\vv_k|\vq_{1,k}|\cdots|\vq_{d-1,k}]$. Since $\vv_k\perp\vq_{i,k}$
    for every $i$, we have $\mM^\top\mM=\diag(\|\vv_k\|_{\ell^2}^2,1,\ldots,1)$ and hence
    $\left|\det\mM\right|=\|\vv_k\|_{\ell^2}$. Applying a linear change of
    variables and Tonelli's theorem,
    \begin{align*}
        \|u_{\vtheta_m}-u_{\vtheta}\|_{\Lone(\Omega,\R^{d'})}
         & =\int_{\R^d}g_m(\vx)\,d\vx
        =\left|\det\mM\right|\int_{\R^d}g_m(\mM\vx)\,d\vx                  \\
         & =\|\vv_k\|_{\ell^2}\int_{H_k}\int_{\R}g_m(\vy+t\vv_k)\,dt\,d\vy
        =\|\vv_k\|_{\ell^2}\int_{H_k}\Upsilon_{m,k}(\vy)\,d\vy,
    \end{align*}
    that is,
    \begin{equation}
        \label{eq:properness_slicing}
        \int_{H_k}\Upsilon_{m,k}(\vy)\,d\vy
        =\frac{1}{\|\vv_k\|_{\ell^2}}
        \|u_{\vtheta_m}-u_{\vtheta}\|_{\Lone(\Omega,\R^{d'})}.
    \end{equation}

    The functions $\Upsilon_{m,k}$ are nonnegative and measurable, so by
    Tonelli's theorem for series, \eqref{eq:properness_slicing} and
    \eqref{eq:properness_summable_error},
    \[
        \int_{H_k}\sum_{m=1}^{\infty}\Upsilon_{m,k}(\vy)\,d\vy
        =\sum_{m=1}^{\infty}\int_{H_k}\Upsilon_{m,k}(\vy)\,d\vy
        \le\frac{1}{\|\vv_k\|_{\ell^2}}\sum_{m=1}^{\infty}2^{-m}
        =\frac{1}{\|\vv_k\|_{\ell^2}}<\infty.
    \]
    Therefore, the set
    \begin{equation}
        \label{eq:properness_nullset}
        N_k=\Bigl\{\vy\in H_k:\sum_{m=1}^{\infty}\Upsilon_{m,k}(\vy)=+\infty\Bigr\}
    \end{equation}
    has zero $(d-1)$-dimensional Lebesgue measure, and for every
    $\vy\in H_k\setminus N_k$ the series $\sum_m\Upsilon_{m,k}(\vy)$ converges,
    so that its nonnegative summands satisfy $\Upsilon_{m,k}(\vy)\to0$ as
    $m\to\infty$.

    The conclusion just obtained is vacuous for those $\vy$ whose slice
    $\Omega_{\vy,k}$ is empty, so we must select $\vy\notin N_k$ whose slice
    is not. Applying
    the same change of variables to $\mathbf 1_{\Omega}$ in place of $g_m$ gives
    \[
        \int_{H_k}\calL^1(\Omega_{\vy,k})\,d\vy
        =\frac{1}{\|\vv_k\|_{\ell^2}}\calL^d(\Omega)>0,
    \]
    since $\Omega$ is nonempty and open. Hence, the set
    \[
        S_k=\{\vy\in H_k:\calL^1(\Omega_{\vy,k})>0\},
    \]
    which is measurable because $\vy\mapsto\calL^1(\Omega_{\vy,k})$ is (again by
    Tonelli's theorem), has positive $(d-1)$-dimensional measure.
    Since $\calL^{d-1}(N_k)=0<\calL^{d-1}(S_k)$, the set $S_k$ is not contained
    in $N_k$, and we may choose
    \[
        \vy_k\in S_k\setminus N_k,
        \qquad\text{so that}\qquad
        \calL^1(\Omega_{\vy_k,k})>0
        \quad\text{and}\quad
        \sum_{m=1}^{\infty}\Upsilon_{m,k}(\vy_k)<+\infty.
    \]
    The slice $\Omega_{\vy_k,k}$ is then open and nonempty, having positive
    measure, so it contains a nonempty bounded open interval $I_k$ with
    $\overline{I_k}\subset\Omega_{\vy_k,k}$.

    \textit{Step 3: boundedness of the parameters.}
    On $I_k$, the restriction of the target along the chosen line is
    \[
        \begin{split}
            v^k_{\vtheta}(t)
             & =u_{\vtheta}(\vy_k+t\vv_k)
            =\vs+\sum_{i=1}^{n}\tanh\bigl(\vw_i^\top(\vy_k+t\vv_k)+b_i\bigr)\va_i \\
             & =\vs+\sum_{i=1}^{n}\tanh(\alpha_it+\beta_i)\va_i,
        \end{split}
    \]
    where $\alpha_i\coloneqq\vw_i^\top\vv_k$ and
    $\beta_i\coloneqq\vw_i^\top\vy_k+b_i$. The function $v^k_{\vtheta}$ is
    a shallow tanh neural network with hidden dimension $n$, input dimension $1$ and the
    same amplitudes $\va_i$ and offset $\vs$ as $u_{\vtheta}$. By
    \Cref{lem:generic_directions} its parameters satisfy $\alpha_i\neq0$ for
    every $i$ and $(\alpha_i,\beta_i)\neq\pm(\alpha_j,\beta_j)$ for $i\neq j$,
    while $\va_i\neq\mathbf 0$ since $\vtheta$ is irreducible; hence
    $v^k_{\vtheta}$ is irreducible, as observed in
    \Cref{rem:generic_directions_scope}. The restrictions
    $v^k_{\vtheta_m}(t)=u_{\vtheta_m}(\vy_k+t\vv_k)$ of the approximating
    networks have parameters
    \begin{equation}
        \label{eq:properness_restricted_params}
        \bigl(\vw_{j,m}^\top\vv_k,
        \ \vw_{j,m}^\top\vy_k+b_{j,m},\ \va_{j,m},\ \vs_m\bigr),
        \qquad 1\le j\le n.
    \end{equation}
    Note that those parameters need not be irreducible to apply
    \Cref{lem:properness_interval}. Let then $I_k\subseteq\Omega_{\vy_k,k}$
    be an interval as in the previous step of this proof. It holds that
    \[
        \|v^k_{\vtheta_m}-v^k_{\vtheta}\|_{\Lone(I_k,\R^{d'})}
        =\int_{I_k}\|u_{\vtheta_m}(\vy_k+t\vv_k)-u_{\vtheta}(\vy_k+t\vv_k)\|_{\ell^1}\,dt
        \le\Upsilon_{m,k}(\vy_k)\xrightarrow[m\to\infty]{}0 .
    \]
    Therefore, \Cref{lem:properness_interval}, applied on the bounded open interval $I_k$,
    bounds the subsequence of parameters
    \eqref{eq:properness_restricted_params} uniformly in $m$.

    Repeating the construction above for every $k=1,\ldots,d$, we obtain that
    the parameters \eqref{eq:properness_restricted_params} are bounded uniformly
    in $m$ for every $k$. With $\mV$ as in \eqref{eq:properness_basis_matrix} we
    then obtain, for every $1\le j\le n$,
    \[
        \|\vw_{j,m}\|_{\ell^2}
        =\|\mV^{-\top}\mV^\top\vw_{j,m}\|_{\ell^2}
        \le\|\mV^{-\top}\|_{\ell^2}\,\|\mV^\top\vw_{j,m}\|_{\ell^2}
        =\|\mV^{-\top}\|_{\ell^2}
        \left(\sum_{k=1}^{d}|\vw_{j,m}^\top\vv_k|^2\right)^{1/2},
    \]
    which is bounded uniformly in $m$ since $\{\vw_{j,m}^\top\vv_k\}_m$ is
    bounded for each $k$. The biases are then bounded by the triangle inequality
    \[
        |b_{j,m}|\le
        |\vw_{j,m}^\top\vy_1+b_{j,m}|
        +\|\vy_1\|_{\ell^2}\,\|\vw_{j,m}\|_{\ell^2},
    \]
    while the amplitudes $\va_{j,m}$ and the offset $\vs_m$ were already bounded
    above. Hence the whole parameter sequence
    $\{\vw_{j,m},b_{j,m},\va_{j,m},\vs_m, \, 1\le j\le n\}_m$ is bounded,
    contradicting $\|\vtheta_m\|\to+\infty$. This proves that $\{\vtheta_m\}$ is
    bounded in $\R^N$, with no assumption on the connectedness or on the
    boundary of $\Omega$.
\end{proof}

\begin{corollary}
    \label{cor:limit_points}
    Under the assumptions of \Cref{thm:boundedness}, every limit point
    $\hat\vtheta$ of $\{\vtheta_m\}$, which exists by
    \Cref{thm:boundedness}, belongs to $\mTheta$ and satisfies
    $\hat\vtheta \sim \vtheta$.
\end{corollary}

\begin{proof}
    Let $\hat\vtheta$ be a limit point of $\{\vtheta_m\}$, say
    $\vtheta_{m_l}\to\hat\vtheta$. The realization map is continuous from $\R^N$
    into $\Lp(\Omega,\R^{d'})$, hence $u_{ \hat\vtheta } = u_{\vtheta}$ in
    $\rmL^p$ and everywhere in $\Omega$ by continuity. Since $\vtheta\in
        \mTheta$, by \Cref{cor:not_equal_lower_dim} also
    $\hat\vtheta\in\mTheta$, and finally \Cref{cor:non_injectivity} gives
    $\hat\vtheta\sim\vtheta$.
\end{proof}

\section{Proof of the manifold structure}
\label{sec:proof}
In this section, we prove \Cref{thm:main}. We first compute the
differential of the realization map (\Cref{lem:partial_realization})
and show that it is injective at irreducible parameters
(\Cref{lem:injective_differential}). Together with \Cref{thm:boundedness} and
\Cref{cor:limit_points}, which give the continuity of the inverse, this
implies the main result. Finally, we prove \Cref{thm:connectedness} on the connected components of irreducible networks.
\begin{restatable}{lemma}{partialparam}
    \label{lem:partial_realization}
    Let $\Omega \subset \R^d$ be an open and bounded set, let $p \in [1,
            +\infty]$, and let $\ell\in\N$. The realization map $\rho: \R^N \to
        \rmW^{\ell,p}(\Omega, \R^{d'})$ is a smooth map. Moreover, its
    differential at $\vtheta \in \R^N$ is given by
    \[
        \D\rho(\vtheta)[\vh] = \sum_{i=1}^{N} \partial_{\theta_i} u_{\vtheta} \, \cdot h_i,
    \]
    where the partial derivatives $\partial_{\theta_i} u_{\vtheta}$ are given
    by:
    \begin{align}
        \partial_{\mW_{ij}} u_{\vtheta}(\vx) & =  \tanh'(\vw_i^\top \vx + b_i)\, x_j \cdot \va_i, \\
        \partial_{b_{i}} u_{\vtheta}(\vx)    & =  \tanh'(\vw_i^\top \vx + b_i) \cdot \va_i,       \\
        \partial_{\mA_{ki}} u_{\vtheta}(\vx) & =  \tanh(\vw_i^\top \vx + b_i) \cdot \ve_k,        \\
        \partial_{s_{k}} u_{\vtheta}(\vx)    & =  \ve_k.
    \end{align}
\end{restatable}
The formulas above are clearly the pointwise partial derivatives of
$u_{\vtheta}(\vx)$ with respect to the parameters, in the classical sense.
One has to show that they are also the Fr\'echet derivatives of $\rho$
as a map with values in $\rmW^{\ell,p}(\Omega,\R^{d'})$, and that $\rho$ is
smooth; we postpone the proof to \Cref{sec:apx_manifold_proofs}.

The following lemma on the injectivity of the differential at irreducible
parameters is closely related to a classical result of Fukumizu
\cite[Theorem~1]{fukumizu_regularity_1996}, from which it can also be deduced.
We include the short proof below since it follows directly from
\Cref{lem:generalized_independence}.

\begin{lemma}
    \label{lem:injective_differential}
    With the same assumptions and notation as in
    \Cref{lem:partial_realization}, the differential $\D\rho(\vtheta)$ at an
    irreducible parameter $\vtheta\in\mTheta$ is injective.
\end{lemma}
\begin{proof}
    Let $\tilde{\vtheta} \in \R^N$ be such that
    $\D\rho(\vtheta)[\tilde{\vtheta}] = 0$, i.e.,
    \begin{multline*}
        \D\rho(\vtheta)[\tilde{\vtheta}] = \sum_{i,j} {\tilde{\mW}_{ij}} \cdot \tanh'(\vw_i^\top \vx + b_i)\, x_j \cdot \va_i + \sum_{i} {\tilde{b}_{i}}\cdot \tanh'(\vw_i^\top \vx + b_i) \cdot \va_i + \\
        + \sum_{i,k} {\tilde{\mA}_{ki}}\cdot \tanh(\vw_i^\top \vx + b_i) \cdot \ve_k + \sum_{k} {\tilde{s}_{k}} \cdot \ve_k = 0
        \qquad \text{for all } \vx \in \Omega.
    \end{multline*}
    In each output coordinate $k$, this is a vanishing linear combination on
    $\Omega$ of the functions $1$, $\tanh(\vw_i^\top \vx + b_i)$,
    $\tanh'(\vw_i^\top \vx + b_i)$ and $x_j \tanh'(\vw_i^\top \vx + b_i)$,
    with coefficients $\tilde{s}_k$, $\tilde{\mA}_{ki}$,
    $\tilde{b}_i [\va_i]_k$ and $\tilde{\mW}_{ij} [\va_i]_k$, respectively.
    These functions belong to the family of
    \Cref{lem:generalized_independence}, applied with $U = \Omega$, $J = 1$,
    $A = 1$ and $S = 0$, whose hypotheses hold because $\vtheta$ is
    irreducible: $\vw_i \neq \mathbf 0$ and the pairs $(\vw_i, b_i)$ are
    pairwise distinct up to sign. Thus, all the coefficients vanish, that is,
    $\tilde{s}_k = 0$ and $\tilde{\mA}_{ki} = 0$ for all $i, k$, and
    $\tilde{\mW}_{ij} [\va_i]_k = \tilde{b}_i [\va_i]_k = 0$ for all
    $i, j, k$. For every $i$, choosing an output coordinate $k$ with
    $[\va_{i}]_k \neq 0$, which exists by irreducibility of $\vtheta$, gives
    $\tilde{\mW}_{ij} = 0$ for all $j$ and $\tilde{b}_{i} = 0$. Hence,
    $\tilde{\vtheta} = 0$, proving the injectivity of the differential.
\end{proof}

With these ingredients, we can now prove \Cref{thm:main}, which we restate
for convenience.

\mainthm*

\begin{proof}
    We aim to prove that the set $\M$ of irreducible shallow tanh neural
    networks is a smooth embedded submanifold of $\rmW^{\ell,p}(\Omega,
        \R^{d'})$ of dimension $N = n d + n + d' n + d'$. The set $\M$ is precisely the
    image of $\tilde{\rho}$ (recall that $\tilde{\rho}$ is the unique
    realization acting on equivalence classes such that $\tilde{\rho}([\vtheta])
        = \rho(\vtheta)$). We aim to prove that $\tilde{\rho}$ is a smooth
    injective embedding so that by
    \cite[Theorem H and Lemma 1.13]{glockner_fundamentals_2016}
    the image is a smooth embedded submanifold of $\rmV$. Since $\mTheta / G$ is
    finite dimensional, by
    \cite[Theorem H and Lemma 1.13]{glockner_fundamentals_2016}
    it suffices to show that:
    \begin{enumerate}[(i)]
        \item \label{def:item:proof_1} $\D\tilde{\rho}([\vtheta])$ is injective $\forall [\vtheta]
                  \in \mTheta / G$;
        \item \label{def:item:proof_2} $\tilde{\rho}$ is a topological embedding, i.e.,
              the corestriction $\tilde{\rho}_{\M}:\mTheta/G\to\M$,
              $\tilde{\rho}_{\M}([\vtheta])=\tilde{\rho}([\vtheta])$,
              has a continuous inverse $\tilde{\rho}_{\M}^{-1}:\M\to\mTheta/G$.
    \end{enumerate}

    For (\ref{def:item:proof_1}), by \Cref{lem:injective_differential} we know
    that for all $\vtheta\in\mTheta$
    \[
        \D\rho(\vtheta) = \D\tilde{\rho}([\vtheta]) \circ \D\pi(\vtheta)
    \]
    is injective. Since $\D\pi(\vtheta)$ is surjective, $\D\tilde{\rho}([\vtheta])$ is injective as well.

    To prove (\ref{def:item:proof_2}), it is sufficient to prove sequential
    continuity of the inverse $\tilde{\rho}_{\M}^{-1}$. Its domain $\M\subset\rmV$ has the metric
    induced by the norm of $\rmV$, and its codomain has the metric $d_G$.
    Let $\{u_m =
        \tilde{\rho}([\vtheta_m])\}_{m\in\N}\subset\M$ and $u =
        \tilde{\rho}([\vtheta])$ be such that $u_m \xrightarrow{\rmV} u$. We
    aim to show that $[\vtheta_m] \xrightarrow{\mTheta/G} [\vtheta]$; by
    \Cref{lem:quotient_metric}, this amounts to $d_G([\vtheta_m], [\vtheta])
        \to 0$. Suppose not: then there exist $\varepsilon > 0$ and a
    subsequence with $d_G([\vtheta_{m_k}], [\vtheta]) \ge \varepsilon$ for all
    $k$. Since $u_{\vtheta_{m_k}} = \rho(\vtheta_{m_k}) \to u = \rho(\vtheta)$
    in $\rmV$, hence in $\Lp(\Omega, \R^{d'})$, and $\vtheta \in \mTheta$ is irreducible,
    \Cref{thm:boundedness} guarantees that $\{\vtheta_{m_k}\}_{k \in \N}$ is
    bounded in $\R^N$; a further subsequence therefore converges,
    $\vtheta_{m_{k_j}} \to \hat{\vtheta}$. By
    \Cref{cor:limit_points}, $\hat\vtheta \in \mTheta$ and $\hat\vtheta \sim
        \vtheta$, i.e., $[\hat\vtheta] = [\vtheta]$. But then
    \[
        d_G\bigl([\vtheta_{m_{k_j}}], [\vtheta]\bigr)
        = d_G\bigl([\vtheta_{m_{k_j}}], [\hat\vtheta]\bigr)
        \le \normsmall{\vtheta_{m_{k_j}} - \hat\vtheta} \longrightarrow 0,
    \]
    contradicting $d_G([\vtheta_{m_k}], [\vtheta]) \ge \varepsilon$. This
    proves (\ref{def:item:proof_2}) and concludes the proof.
\end{proof}

We now prove \Cref{thm:connectedness}, which we restate for convenience.

\connectedness*
\begin{proof}

    By definition, $\mTheta$ is obtained by removing from $\R^N$ the linear
    subspaces $\{\vtheta \in \R^N : \vw_i = \mathbf{0}_d\}$, of codimension
    $d$, $\{\vtheta \in \R^N : \va_i = \mathbf{0}_{d'}\}$, of codimension
    $d'$, and $L_{ij}^\pm \coloneqq \{\vtheta \in \R^N : (\vw_i, b_i) = \pm(\vw_j,
        b_j)\}$, $i<j$, of codimension $d+1 \ge 2$. Among these, the only
    hyperplanes are the subspaces $\{\vw_i = \mathbf 0\}$ when $d = 1$ and
    $\{\va_i = \mathbf 0\}$ when $d' = 1$.

    \emph{Proof of (i).} Let $\max(d, d') \ge 2$ and set
    \[
        U \coloneqq
        \begin{cases}
            \{w_i > 0\ \forall i\} & \text{if } d = 1,\ d' \ge 2, \\
            \{a_i > 0\ \forall i\} & \text{if } d \ge 2,\ d' = 1, \\
            \R^N                   & \text{if } d, d' \ge 2.
        \end{cases}
    \]
    For every $\vtheta \in \mTheta$, the element $(\vepsilon, \mathrm{id}) \in G$
    with
    \[
        \epsilon_i \coloneqq
        \begin{cases}
            \operatorname{sign} w_i & \text{if } d = 1,\ d' \ge 2, \\
            \operatorname{sign} a_i & \text{if } d \ge 2,\ d' = 1, \\
            1                       & \text{if } d, d' \ge 2,
        \end{cases}
    \]
    maps $\vtheta$ into $U$.
    Hence, $P \coloneqq \mTheta \cap U$ has nonempty intersection with every $G$-orbit, and $\M =
        \rho(P)$. The open convex set $U$ does not intersect the hyperplanes listed
    above, so $P$ is $U$ minus finitely many linear subspaces of codimension
    at least $2$. Therefore, $P$ is connected
    \cite[Lemma~A.1]{schmeding_diffeomorphism_2015}, and so is $\M =
        \rho(P)$, since $\rho$ is continuous.

    \emph{Proof of (ii).} Let $d = d' = 1$, $\kappa(\vtheta) \coloneqq \#\{i : w_i
        \cdot a_i > 0\}$ and $\mTheta_k \coloneqq \{\vtheta \in \mTheta :
        \kappa(\vtheta) = k\}$, so that $\M_k = \rho(\mTheta_k)$.
    The sets
    $\{\mTheta_k\}_{k\in \{0, \dots, n\}}$ are open in the topology of
    $\R^N$ and they partition $\mTheta$.
    Each $\mTheta_k$ is also $G$-invariant, since $G$ permutes
    the neurons and changes the signs of $w_i$ and $a_i$ simultaneously.
    Hence $\pi^{-1}(\pi(\mTheta_k)) = \mTheta_k$ is open, so $\pi(\mTheta_k)$
    is open in the quotient topology of $\mTheta/G$. Moreover, the sets
    $\pi(\mTheta_k)$ are pairwise disjoint and cover $\mTheta/G$. Each
    $\pi(\mTheta_k)$ is therefore also closed, since its complement
    $\bigcup_{j \neq k} \pi(\mTheta_j)$ is a finite union of open sets.
    Since $\tilde\rho_{\M}$ is a homeomorphism, the sets $\M_k =
        \tilde\rho_{\M}(\pi(\mTheta_k))$ form a partition of $\M$ into clopen
    subsets.

    Let $U_k \coloneqq \{\vtheta\in\R^N : w_i > 0\ \forall i,\ a_i > 0 \text{ for } i \le k,\ a_i < 0
        \text{ for } i > k\}$
    be the set of parameters whose weights
    $w_i$ are all positive and for which $w_i a_i > 0$ exactly for the first
    $k$ neurons.
    Define $P_k \coloneqq \mTheta \cap U_k \subseteq \mTheta_k$. Given
    $\vtheta \in \mTheta_k$, changing the signs of the neurons with $w_i < 0$
    yields a parameter with $w_i > 0$ for all $i$ and, by invariance of
    $\kappa$, exactly $k$ indices with $a_i > 0$; a permutation then moves
    these neurons to the first $k$ positions. Hence, $P_k$ intersects every
    $G$-orbit in $\mTheta_k$, and $\M_k = \rho(P_k)$. Since $U_k$ is
    nonempty, open, convex, and disjoint from $\{w_i = 0\}$, $\{a_i = 0\}$
    and $L_{ij}^-$, we have $P_k = U_k \setminus \bigcup_{i<j} L_{ij}^+$,
    which is nonempty and connected because each $L_{ij}^+$ has codimension
    $2$. Thus, each $\M_k$ is a nonempty, connected, clopen subset of $\M$,
    that is, a connected component of $\M$ \cite[Section~32, Theorem~C]{Simmons1963}. Hence, $\M_0, \dots, \M_n$ are
    exactly the $n+1$ connected components of $\M$.
\end{proof}

\section{Conclusion}
\label{sec:conclusion}
We proved that the set $\M$ of irreducible shallow $\tanh$ neural networks with
fixed input dimension $d$, output dimension $d'$ and hidden dimension $n$ is a
smooth embedded submanifold of $\rmW^{\ell,p}(\Omega,\R^{d'})$, for every
nonempty open and bounded set $\Omega\subset\R^d$, every $\ell\in\N$ and every
$p\in[1,+\infty]$ (\Cref{thm:main}). The key step of the proof is \Cref{thm:boundedness}:
if a sequence of shallow $\tanh$ networks converges in $\Lp(\Omega)$ to an
irreducible network with the same number of neurons, then its parameters remain
bounded. Moreover, we proved that $\M$ is connected unless
$d=d'=1$, in which case it has $n+1$ connected components
(\Cref{thm:connectedness}).

Several directions remain open. The first is the extension to other activation
functions. Our results do not extend to the ReLU
\cite{elisenda_grigsby_functional_2025}, whereas they carry over directly to the
logistic sigmoid (\Cref{rem:sigmoid}).
Since our
proofs rely on specific properties of the hyperbolic tangent, other activation
functions require a case-by-case analysis. Second, the extension of our results to deep
networks
requires the characterization of irreducible parameters of
\cite{fefferman1994,vlacic_affine_2021,vlacic_neural_2022}. Developing
the deep counterparts of \Cref{thm:boundedness} and
\Cref{lem:injective_differential} is the subject of ongoing work. Finally,
as mentioned in the introduction, our work constitutes the basis for a
theoretical analysis of Riemannian optimization algorithms for neural networks.

\section*{Declaration on the use of AI}
The authors used ChatGPT-6 Astra (OpenAI) and Claude Opus~5.5 (Anthropic) to
assist in polishing the manuscript and in developing some of the proofs. In
particular, the strategy for the proof of \Cref{thm:boundedness} was devised by
the authors, and ChatGPT-6 Astra was used to develop some of the arguments; the
authors then refined and polished the proof proposed by the model. In
\Cref{thm:connectedness}, Claude Opus~5.5 was used to fully characterize the
connected components in the case of scalar input and output. All other proofs
were derived entirely by the authors. Claude Opus~5.5 was used to improve the
writing (grammar and spelling).

    {\small
        \bibliographystyle{plain}
        \bibliography{journals_full,bibliography}}

\appendix
\crefalias{section}{appendix}
\section{Proof of generalized linear independence}
\label{sec:lin_indep}
In this appendix we give a detailed proof of
\Cref{lem:generalized_independence}. We first prove it in one dimension
(\Cref{lem:generalized_independence_1d}), and then deduce the multi-dimensional
case by
restricting to a suitably chosen line.
The proof of the following statement is based on the
technique of \cite{fefferman1994} and follows closely \cite[Section
    3.1]{vlacic_affine_2021}. We present it here fully, as we need the result in a
slightly more general setting than the one in the cited references.

\begin{lemma}[Generalized linear independence: one-dimensional case]
    \label{lem:generalized_independence_1d}
    Let $I \subseteq \R$ be a nonempty open interval. Let $(\alpha_1, \beta_1),
        \dots, (\alpha_R, \beta_R) \in (\R \setminus \{0\}) \times \R$ be
    pairwise distinct up to sign, i.e., $(\alpha_r, \beta_r) \neq
        \pm(\alpha_{r'}, \beta_{r'})$ for $r \neq r'$; let $\lambda_1, \dots,
        \lambda_S \in \R \setminus \{0\}$ be pairwise distinct; let $J \in \N$;
    and let $p_0$, $p_{r,j}$ ($1 \le r \le R$, $0 \le j \le J$) and $q_s$ ($1
        \le s \le S$) be real polynomials. If
    \begin{equation}
        \label{eq:gen_indep_identity_1d}
        p_0(t) + \sum_{r=1}^{R} \sum_{j=0}^{J} p_{r,j}(t)\,
        \tanh^{(j)}(\alpha_r t + \beta_r) + \sum_{s=1}^{S} q_s(t)\,
        e^{\lambda_s t} = 0 \qquad \text{for all } t \in I,
    \end{equation}
    then all the polynomials $p_0$, $p_{r,j}$, $q_s$ vanish identically.
\end{lemma}
\begin{proof}
    Denote the left-hand side of \eqref{eq:gen_indep_identity_1d} by
    \[
        f(t) \coloneqq p_0(t) + \sum_{r=1}^{R} \sum_{j=0}^{J} p_{r,j}(t)\,
        \tanh^{(j)}(\alpha_r t + \beta_r) + \sum_{s=1}^{S} q_s(t)\,
        e^{\lambda_s t}.
    \]
    The function $f$ extends to a meromorphic function on $\C$, with poles
    contained in the pole sets of the functions $t \mapsto \tanh^{(j)}(\alpha_r
        t + \beta_r)$. Since this extension vanishes on a real interval, and its
    domain (the plane minus a discrete set) is open and connected, the identity
    theorem \cite{cartan_elementary_1963} shows that $f$ vanishes identically on
    its domain. In particular, the poles of $f$ are removable, i.e., all its
    principal parts vanish.

    \emph{Step 1: the $\tanh$ coefficients vanish.} Using the parity relation
    \[
        \tanh^{(j)}(-\zeta) = (-1)^{j+1}\tanh^{(j)}(\zeta),
    \]
    we may replace
    $(\alpha_r, \beta_r)$ by $-(\alpha_r, \beta_r)$ where necessary, absorbing
    the signs into the coefficient polynomials, so that we may assume in the
    following that $\alpha_r > 0$ for all $r$ and that all pairs
    $(\alpha_r, \beta_r)$ are distinct.
    The poles of $t \mapsto
        \tanh(\alpha_{r} t + \beta_{r})$ form the lattice
    \[
        z_{r,k} = \frac{-\beta_{r} + i\pi\left(k + \tfrac12\right)}{\alpha_{r}},
        \qquad k \in \Z,
    \]
    contained in the vertical line $\Re z = -\beta_{r}/\alpha_{r}$, with spacing
    $\pi/\alpha_{r}$.


    Let
    \begin{equation*}
        \mathcal{R} \coloneqq \{r \in \{1, \dots, R\}: \exists j \in \{0, \dots, J\} \text{ such that } p_{r,j}\text{ does not vanish}\}.
    \end{equation*}
    Suppose that $\mathcal{R}$ is not empty and fix $\rstar$ such that
    $  \alpha_{\rstar} = \max_{r\in\mathcal{R}}\alpha_r$.
    We claim that infinitely many poles $z_{\rstar,k}$ are
    \emph{private}, i.e., not poles of any other remaining neuron $r\in\calR$.
    If $\mathcal{R} = \{\rstar\}$, this is trivial; otherwise, fix a remaining $\mathcal{R}\ni r \neq \rstar$. If $-\beta_r/\alpha_r \neq
        -\beta_{\rstar}/\alpha_{\rstar}$, the two pole lattices lie on different vertical
    lines and share no point; this covers in particular the case $\alpha_r
        = \alpha_{\rstar}$, in which $\beta_r \neq \beta_{\rstar}$. Otherwise
    $\alpha_r < \alpha_{\rstar}$ by definition of $\rstar$, and, comparing
    imaginary parts, $z_{r, k'} = z_{\rstar, k}$ for some $k, k' \in \Z$ holds
    if and only if $\alpha_r(2k+1) = \alpha_{\rstar}(2k'+1)$, i.e.,
    \begin{equation*}
        \alpha_r/\alpha_{\rstar} = (2k'+1)/(2k+1) \in \mathbb{Q}.
    \end{equation*}
    Thus, if
    $\alpha_r/\alpha_{\rstar}$ is irrational, the two lattices share no point at
    all. If instead $\alpha_r/\alpha_{\rstar}$ is rational, write
    $\alpha_r/\alpha_{\rstar} = p_r/q_r$ in lowest terms, $1 \le p_r <
        q_r$; the common
    pole condition becomes $p_r(2k+1) = q_r(2k'+1)$, which
    implies $q_r \mid (2k+1)$ since $q_r\mid p_r(2k+1)$ and $p_r$
    and $q_r$ are coprime.
    Let then
    \begin{equation*}
        K = \left\{  n Q : n\in \N \text{ with }Q = \prod_{r\neq r_\star: \alpha_r/\alpha_{r_\star}\in \mathbb{Q}} q_r \right\}.
    \end{equation*}
    For all $k\in K$ it holds that
    $2k+1 \equiv 1 \text{ mod } q_r$ for all such $r$, and $q_r \ge 2$ since
    $1 \le p_r < q_r$. Thus none of the divisibility conditions
    $q_r \mid (2k+1)$ can hold, i.e., for $k \in K$ the point $z_{\rstar, k}$
    is not a pole of $\tanh(\alpha_r z + \beta_r)$ for any
    $r \in \calR \setminus \{\rstar\}$. We obtained infinitely many private
    poles $(z_{\rstar,k})_{k\in K}$.

    Let $\jstar$ be the largest derivative order with
    $p_{\rstar,\jstar} \not\equiv 0$.
    Near a pole $\zeta_0$ of $\tanh$ we have $\tanh(\zeta) = (\zeta -
        \zeta_0)^{-1} + O(1)$, hence, near $z_{r,k}$,
    \[
        \tanh^{(j)}(\alpha_r z + \beta_r)
        = \frac{(-1)^j j!\, \alpha_r^{-(j+1)}}{(z - z_{r,k})^{j+1}}
        + \text{terms with poles of lower order} .
    \]
    At a private pole $z_{\rstar,k}$ every summand of
    \eqref{eq:gen_indep_identity_1d} not involving $\rstar$ is holomorphic,
    the terms $p_{\rstar,j}\tanh^{(j)}(\alpha_{\rstar}z+\beta_{\rstar})$ with
    $j<\jstar$ have poles of order at most $\jstar$, and multiplication by a
    polynomial does not increase the order of a pole. Hence the coefficient
    of $(z - z_{\rstar,k})^{-(\jstar+1)}$ in the Laurent expansion of $f$
    equals $(-1)^{\jstar} \jstar!\,
        \alpha_{\rstar}^{-(\jstar+1)}\,p_{\rstar,\jstar}(z_{{\rstar},k})$. Since all principal parts
    vanish, $p_{{\rstar},\jstar}(z_{{\rstar},k}) = 0$ at infinitely many distinct
    points, so
    $p_{\rstar,\jstar} \equiv 0$, a contradiction. It follows that $\mathcal{R}$
    is empty, i.e., all the $\tanh$ coefficients $p_{r,j}$ vanish.
    
    \emph{Step 2: the polynomial and exponential coefficients vanish.} All
    the poles have nonzero imaginary part, so $\R$ is contained in the domain
    of $f$, on which $f$ vanishes identically. Since all the $p_{r,j}$ vanish,
    the identity \eqref{eq:gen_indep_identity_1d} is therefore reduced to
    \begin{equation}
        \label{eq:poly_exp_eqzero}
        \sum_{s=0}^{S} q_s(t)\, e^{\lambda_s t} = 0 \qquad \text{for all } t \in \R,
    \end{equation}
    where we have set $q_0 := p_0$ and $\lambda_0 := 0$; by assumption the rates
    $\lambda_0, \dots, \lambda_S$ are pairwise distinct. Suppose by
    contradiction that $\lambda_{\star } \coloneqq \lambda_{s_\star}$ is the
    largest $\lambda_s$ such that the associated $q_s$ does not vanish, and let
    $q_\star$ be the associated polynomial. Then
    \begin{equation*}
        q_\star(t) = -\sum_{s\neq s_\star} q_s(t) e^{( \lambda_s - \lambda_\star )t} \to 0 \qquad \text{as }t\to \infty.
    \end{equation*}
    This limit can hold only if $q_\star \equiv 0$, which is a
    contradiction.
\end{proof}

We now prove \Cref{lem:generalized_independence}, which we restate for convenience.
\genindep*

\begin{proof}
    Linear independence of the family is equivalent to the following claim: if
    \begin{equation}
        \label{eq:gen_indep_identity}
        P_0(\vx) + \sum_{r=1}^{R} \sum_{j=0}^{J} P_{r,j}(\vx)\,
        \tanh^{(j)}(\vw_r^\top \vx + b_r) + \sum_{s=1}^{S} Q_s(\vx)\,
        e^{\vmu_s^\top \vx} = 0 \qquad \text{for all } \vx \in U,
    \end{equation}
    where $P_0$, $P_{r,j}$, $Q_s$ are real polynomials of degree at most $A$,
    then all these polynomials vanish identically. We will refer to those
    polynomials as ``coefficient polynomials'' throughout the rest of the proof.
    We prove the claim by
    reducing it, through restriction to a generic line, to the
    one-dimensional case of \Cref{lem:generalized_independence_1d}.

    Assume by contradiction that at least one coefficient polynomial is
    nonzero. Since the zero set of a nonzero polynomial has empty interior, we
    may choose $\vx_* \in U$ at which every nonzero coefficient polynomial
    occurring in \eqref{eq:gen_indep_identity} is nonzero. Next, choose a
    direction $\vu \in \R^d$ so that
    \[
        \begin{gathered}
            \vw_r^\top \vu \neq 0 \quad \forall r, \qquad
            \vw_r^\top \vu \neq \pm\, \vw_{r'}^\top \vu \, \text{ if } \vw_r \neq \pm \vw_{r'}, \\
            \vmu_s^\top \vu \neq 0 \quad \forall s, \qquad
            \vmu_s^\top \vu \neq \vmu_{s'}^\top \vu \quad \forall s \neq s'.
        \end{gathered}
    \]
    It is possible to find such a $\vu$ because the hyperplanes defined by the
    vanishing of each of these linear forms are proper subspaces of $\R^d$, and
    there are only finitely many of them.
    Since $U$ is open, there is a nonempty open interval $I \ni 0$ such that
    $\vx_* + t\vu \in U$ for all $t \in I$. Restricting
    \eqref{eq:gen_indep_identity} to the line $t \mapsto \vx_* + t\vu$ gives
    exactly the one-dimensional identity \eqref{eq:gen_indep_identity_1d} on
    $I$, with
    \[
        \alpha_r := \vw_r^\top \vu, \qquad \beta_r := \vw_r^\top \vx_* + b_r,
        \qquad \lambda_s := \vmu_s^\top \vu,
    \]
    and with polynomial coefficients
    \[
        p_0(t) := P_0(\vx_* + t\vu), \qquad
        p_{r,j}(t) := P_{r,j}(\vx_* + t\vu), \qquad
        q_s(t) := Q_s(\vx_* + t\vu)\, e^{\vmu_s^\top \vx_*} .
    \]
    We now check the hypotheses of \Cref{lem:generalized_independence_1d}. By
    the choice of $\vu$ we have $\alpha_r \neq 0$ for all $r$, and the rates
    $\lambda_s$ are pairwise distinct and nonzero (here the assumption
    $\vmu_s \neq \mathbf 0$ is used). The pairs $(\alpha_r, \beta_r)$ are
    pairwise distinct up to sign: if $\vw_r \neq \pm\vw_{r'}$, then $\alpha_r
        \neq \pm\alpha_{r'}$ by the choice of $\vu$; if $\vw_r = \vw_{r'}$ ($r
        \neq r'$), then $b_r \neq b_{r'}$, so $\alpha_r = \alpha_{r'} \neq
        -\alpha_{r'}$ (as $\alpha_r \neq 0$) and $\beta_r \neq \beta_{r'}$;
    and if $\vw_r = -\vw_{r'}$, then $b_r \neq -b_{r'}$, so $\alpha_r =
        -\alpha_{r'} \neq \alpha_{r'}$ and $\beta_r \neq -\beta_{r'}$.

    \Cref{lem:generalized_independence_1d} therefore shows that all the
    polynomials $p_0$, $p_{r,j}$, $q_s$ vanish identically. On the other
    hand, the restriction to the line of every nonzero coefficient polynomial
    of \eqref{eq:gen_indep_identity} is a nonzero polynomial in $t$, because
    its value at $t = 0$ is nonzero by the choice of $\vx_*$: a
    contradiction. Hence all the original polynomials $P_0$, $P_{r,j}$, $Q_s$
    in \eqref{eq:gen_indep_identity} must be zero.
\end{proof}

\section{Proof of the identifiability lemma}
\label{sec:apx_identifiability}
This appendix contains an elementary proof of \Cref{lem:identifiability},
based only on \Cref{lem:generalized_independence}. For the corresponding
results in the literature, see \cite[Theorem 1]{sussmann_uniqueness_1992} and
\cite[Proposition 2]{vlacic_affine_2021}. We restate it here.

\identifiability*

\begin{proof}[Proof of \Cref{lem:identifiability}]
    We write
    $\vw_i,b_i,\va_i, \vs$ for the parameters of $\vtheta$ and
    $\vw'_j,b'_j,\va'_j, \vs'$ for those of $\vtheta'$.

    \emph{Step 1: aligning signs.} Since $\tanh$ is odd, replacing
    $(\vw'_j,b'_j,\va'_j)$ by $-(\vw'_j,b'_j,\va'_j)$ changes neither
    $u_{\vtheta'}$ nor the irreducibility of $\vtheta'$. For each $j$ such that
    $(\vw'_j,b'_j)=-(\vw_i,b_i)$ for some $i$, we perform this flip and set
    $\varepsilon_j:=-1$; otherwise we set $\varepsilon_j:=1$. Denote the
    resulting parameters by
    $(\vw''_j,b''_j,\va''_j):=\varepsilon_j(\vw'_j,b'_j,\va'_j)$. Then every
    pair $(\vw''_j,b''_j)$ is either equal to some $(\vw_i,b_i)$ or distinct up
    to sign from all of them.

    \emph{Step 2: matching.} Let $\mathcal J\subseteq\{1,\dots,n'\}$ be the set
    of indices $j$ with
    \[
        (\vw''_j,b''_j)=(\vw_{\tau(j)},b_{\tau(j)})
    \]
    for some $\tau(j)\in\{1,\dots,n\}$. By irreducibility of $\vtheta$, $\tau(j)$ is
    unique, and by irreducibility of $\vtheta'$ the map
    $\tau:\mathcal J\to\{1,\dots,n\}$ is injective. Hence, for $\vx\in\Omega$,
    \[
        \begin{split}
            0 = u_{\vtheta}(\vx)-u_{\vtheta'}(\vx)
            = \vs-\vs'
             & +\sum_{j\in\mathcal J}\bigl(\va_{\tau(j)}-\va''_j\bigr)
            \tanh(\vw_{\tau(j)}^\top\vx+b_{\tau(j)})                        \\
             & +\sum_{i\notin\tau(\mathcal J)}\va_i\tanh(\vw_i^\top\vx+b_i)
            -\sum_{j\notin\mathcal J}\va''_j\tanh\bigl((\vw''_j)^\top\vx+b''_j\bigr),
        \end{split}
    \]
    and all the pairs appearing on the right-hand side are pairwise distinct
    up to sign, with nonzero weights. Applying
    \Cref{lem:generalized_independence}, with $U=\Omega$ and $J=A=S=0$, in
    each of the $d'$ output coordinates gives $\vs=\vs'$,
    $\va''_j=\va_{\tau(j)}$ for $j\in\mathcal J$, and $\va_i=\mathbf 0$,
    $\va''_j=\mathbf 0$ for the unmatched indices $i\notin\tau(\mathcal J)$,
    $j\notin\mathcal J$. By condition \eqref{def:item:nondeg_3} there are no
    unmatched indices, i.e., $\mathcal J=\{1,\dots,n'\}$ and $\tau$ is a
    bijection onto $\{1,\dots,n\}$. Thus $n'=n$ and
    $(\vw'_j,b'_j,\va'_j)=\varepsilon_j(\vw_{\tau(j)},b_{\tau(j)},\va_{\tau(j)})$
    for every $j$, that is, $\vtheta'=g\cdot\vtheta$ with
    $g=((\varepsilon_j)_{j=1}^n,\tau^{-1})\in G$ (see \Cref{def:action}), and
    hence $\vtheta\sim\vtheta'$.
\end{proof}

\section{Proof of \texorpdfstring{\Cref{lem:partial_realization}}{Lemma}}
\label{sec:apx_manifold_proofs}

This appendix contains the proof of \Cref{lem:partial_realization} used in the proof of
\Cref{thm:main} in \Cref{sec:proof}. We recall it for convenience.
\partialparam*

\begin{proof}[Proof of \Cref{lem:partial_realization}]
    For every spatial multi-index $\alpha \in \N^d$ with $|\alpha| \le \ell$
    and every parameter multi-index $\beta \in \N^N$, differentiating the
    formula of \Cref{def:tanh_nn} shows (by induction on $|\alpha| + |\beta|$)
    that each component of the function
    \[
        (\vtheta, \vx) \longmapsto \partial_{\vtheta}^{\beta} \partial_{\vx}^{\alpha} u_{\vtheta}(\vx)
    \]
    is a finite sum of terms of the form $p(\vtheta, \vx)\,
        \tanh^{(r)}(\vw_i^\top \vx + b_i)$ with $0 \le r \le |\alpha| +
        |\beta|$, plus possibly a polynomial term $p(\vtheta, \vx)$, where each
    $p$ is a polynomial in $(\vtheta, \vx)\in\R^{N+d}$.
    All derivatives of
    $\tanh$ are bounded on the real line, and $\Omega$ is bounded; hence for
    every compact set $K \subset \R^N$ there is a constant $C_{K,\alpha,\beta}$
    such that
    \begin{equation}
        \label{eq:mixed_derivative_bound}
        \sup_{\vtheta \in K}\, \sup_{\vx \in \Omega}
        \bigl| \partial_{\vtheta}^{\beta} \partial_{\vx}^{\alpha} u_{\vtheta}(\vx) \bigr|
        \le C_{K,\alpha,\beta} .
    \end{equation}
    In particular $\partial_{\vtheta}^{\beta} u_{\vtheta} \in
        \rmW^{\ell,p}(\Omega, \R^{d'})$ for every $\vtheta$ and $\beta$: all its
    spatial derivatives up to order $\ell$ are bounded on the bounded set
    $\Omega$.

    Fix $\vtheta_0 \in \R^N$, let $K := \bar B(\vtheta_0, 1)$, and let $q \in
        \N$. For $\vtheta, \vtheta + \vh \in K$, Taylor's theorem in the
    finite-dimensional parameter variable, applied for each fixed $\vx \in
        \Omega$ to the functions $\vtheta \mapsto \partial_{\vx}^{\alpha}
        u_{\vtheta}(\vx)$, with the uniform bound
    \eqref{eq:mixed_derivative_bound} at order $q+1$ controlling the Lagrange
    remainder, gives
    \[
        \Bigl| \partial_{\vx}^{\alpha} u_{\vtheta+\vh}(\vx)
        - \sum_{|\beta| \le q} \frac{\vh^\beta}{\beta!}\,
        \partial_{\vtheta}^{\beta} \partial_{\vx}^{\alpha} u_{\vtheta}(\vx) \Bigr|
        \le C_{K, q, \alpha}\, \norm{\vh}^{q+1}
        \qquad \forall\, \vx \in \Omega .
    \]
    Since $\|f\|_{\Lp(\Omega)} \le |\Omega|^{1/p} \sup_{\Omega} |f|$, with
    $1/p = 0$ when $p = \infty$, taking $\Lp(\Omega)$ norms and summing
    over $|\alpha| \le \ell$ yields
    \begin{equation}
        \label{eq:taylor_wlp}
        \Bigl\| u_{\vtheta+\vh} - \sum_{|\beta| \le q}
        \frac{\vh^\beta}{\beta!}\, \partial_{\vtheta}^{\beta} u_{\vtheta}
        \Bigr\|_{\rmW^{\ell,p}(\Omega,\R^{d'})}
        \le C_{K,q}\, \norm{\vh}^{q+1} .
    \end{equation}
    The case $q = 0$ of \eqref{eq:taylor_wlp}, applied to
    $\partial_{\vtheta}^{\beta} u_{\vtheta}$ in place of $u_{\vtheta}$, shows
    that $\vtheta \mapsto \partial_{\vtheta}^{\beta} u_{\vtheta}$ is
    (Lipschitz-)continuous from $\R^N$ to $\rmW^{\ell,p}(\Omega, \R^{d'})$ for
    every $\beta$. The case $q = 1$ then shows that $\rho$ is Fr\'echet
    differentiable at every $\vtheta$, with
    \[
        \D\rho(\vtheta)[\vh] = \sum_{i=1}^{N} \partial_{\theta_i} u_{\vtheta}\, h_i ,
    \]
    and applying the same argument to the maps $\vtheta \mapsto
        \partial_{\vtheta}^{\beta} u_{\vtheta}$ inductively shows that all
    higher-order derivatives of $\rho$ exist and are continuous. Hence $\rho
        \in C^{\infty}(\R^N, \rmW^{\ell,p}(\Omega, \R^{d'}))$. Finally,
    differentiating $u_{\vtheta}(\vx) = \sum_{i=1}^{n} \tanh(\vw_i^\top \vx +
        b_i)\, \va_i + \vs$ with respect to each parameter yields the formulas
    in the statement.
\end{proof}

\end{document}